\documentclass{clv3}

\usepackage{hyperref}
\usepackage{xcolor}
\definecolor{darkblue}{rgb}{0, 0, 0.5}
\hypersetup{colorlinks=true,citecolor=darkblue, linkcolor=darkblue, urlcolor=darkblue}

\usepackage{booktabs} 
\usepackage{multirow}
\usepackage{amsmath,amsfonts,amssymb}
\usepackage{subcaption}
\usepackage{enumitem}
\usepackage{bm}
\usepackage{comment}
\usepackage{tcolorbox}
\usepackage{setspace} 

\newcommand{\yh}[1]{\textcolor{black}{#1}}
\newcommand{\hq}[1]{\textcolor{black}{#1}}

\newcommand{\hqDec}[1]{\textcolor{black}{#1}}
\newcommand{\GL}[1]{\textcolor{black}{#1}}
\newcommand{\YH}[1]{\textcolor{black}{#1}}

\newcommand{\hqMay}[1]{\textcolor{black}{#1}}

\newcommand{\hqJune}[1]{\textcolor{blue}{#1}}

\runningtitle{\YH{Hierarchical} Interpretation of Neural Text Classification}

\runningauthor{Yan et al.}

\issue{1}{1}{2022}

\pageonefooter{Action editor: Tal Linzen. Submission received: 10 August 2021; revised version received: 17 July 2022; accepted for publication: 3 August 2022.}

\begin{document}

\title{\YH{Hierarchical} Interpretation of Neural Text Classification}

\author{Hanqi Yan$^*$}
\affil{Department of Computer Science\\ University of Warwick, UK\\
\texttt{Hanqi.Yan@warwick.ac.uk}}

\author{Lin Gui\thanks{Equal contribution}}
\affil{Department of Computer Science\\ University of Warwick, UK\\
\texttt{Lin.Gui@warwick.ac.uk}}

\author{Yulan He}
\affil{Department of Computer Science\\ University of Warwick, UK\\
King's College London, UK\\
Alan Turing Institute, UK\\
\texttt{Yulan.He@warwick.ac.uk}}

\maketitle

\begin{abstract}

Recent years have witnessed increasing interests in developing interpretable models in Natural Language Processing (NLP). Most existing models aim at identifying input features such as words or phrases important for model predictions. Neural models developed in NLP however often compose word semantics in a hierarchical manner. 
As such, interpretation by words or phrases only cannot faithfully explain model decisions \hqMay{in text classification}. This paper proposes a novel \YH{Hierarchical} Interpretable Neural Text classifier, called \textsc{\YH{Hint}}, which can automatically generate explanations of model predictions in the form of label-associated topics in a hierarchical manner. Model interpretation is no longer at the word level, but built on topics as the basic semantic unit. Experimental results on both review datasets and news datasets show that our proposed approach achieves text classification results on par with existing state-of-the-art text classifiers, and generates interpretations more faithful to model predictions and better understood by humans than other interpretable neural text classifiers~\footnote{ Our source code can be accessed at \url{https://github.com/hanqi-qi/SINE}}.

\end{abstract}

\section{Introduction}
\label{sec:introduction}

Deep Learning (DL) models have achieved state-of-the-art performance \yh{in many} NLP tasks \cite{DBLP:conf/naacl/DevlinCLT19,DBLP:conf/nips/YangDYCSL19,DBLP:conf/nips/BrownMRSKDNSSAA20,DBLP:conf/acl/Yan0PH20}. 
\yh{Deep neural network containing many layers is usually viewed as a black box which has a limited interpretability.} 
Recently, \yh{the field of explainable AI (XAI) has exploded with various new approaches proposed to address the problem of the lack} of interpretability of deep learning models \cite{DBLP:journals/cacm/Lipton18,DBLP:conf/acl/JacoviG20,DBLP:conf/acl/RibeiroWGS20}. 

\yh{Methods for the interpretation of DL models can be broadly classified into post-hoc interpretation methods and self-explanatory methods. The former typically aims to establish the relationship between the changes in the prediction output and the changes in the input of a DL model in order to identify features important for model decision. }
\GL{For example,~\citet{jawahar2019does} used probing to examine BERT intermediate layers}. \citet{DBLP:conf/acl/AbdouRBBES20} modified input text by linguistic perturbations and observed their impacts on model outputs.~\citet{DBLP:journals/ijcv/SelvarajuCDVPB20} tracked the impact from gradient changes.~\citet{DBLP:conf/emnlp/KimYKY20} erased word tokens from input text by marginalising out the tokens. 
On the other hand, the self-explanatory models are able to generate explanations during model training by `twinning' black-box ML model with transparent modules. 
For example, in parallel to model learning, an addition module is trained to interpret model behaviour and is used to regularise the model for interpretability \cite{alvarez2018towards,rieger2020interpretations}. Such models however usually require expert prior knowledge or annotated data to guide the learning of interpretability modules. \citet{DBLP:conf/emnlp/ChenJ20} proposed to improve the interpretability of neural text classifiers by inserting variational word masks into the classifier after the word embedding layer in order to filter out noisy word-level features. The interpretations generated by their model are only at the word-level and \hqDec{ignore hierarchical semantic compositions in text.} 

\hqMay{We argue that existing word-level or phrase-level interpretations are not sufficient for interpreting text classifier behaviours, as documents tent to exhibit topic and label shifts. It is therefore more desirable to explore hierarchical structures to capture semantic shift in text at different granularity levels~\citep{DBLP:conf/cikm/OHareDBFSGS09,5710933,DBLP:conf/naacl/YangYDHSH16,pmlr-v108-wang20l,10.1162/tacl_a_00261,10.1145/3442381.3450045,DBLP:journals/artmed/GuiH21}.} \GL{Moreover,} simply \yh{establishing the relationship} 
between the \yh{changes in the} input and \yh{the changes in the} output \yh{in a DL model could identify features which are important for predictions, but ignores subtle interactions among input features. Recent approaches 
have been developed to build explanations through detecting feature interactions \cite{singh2019hierarchical,DBLP:conf/acl/ChenZJ20,DBLP:conf/iclr/JinWDXR20,DBLP:journals/tkde/GuiLZXH22}. Nevertheless, they are only able to identify sub-text-spans which are important for model decisions and largely focus on sentence-level classification tasks.} 

\begin{figure}[!t]
    \centering
     \includegraphics[width=0.95\textwidth]{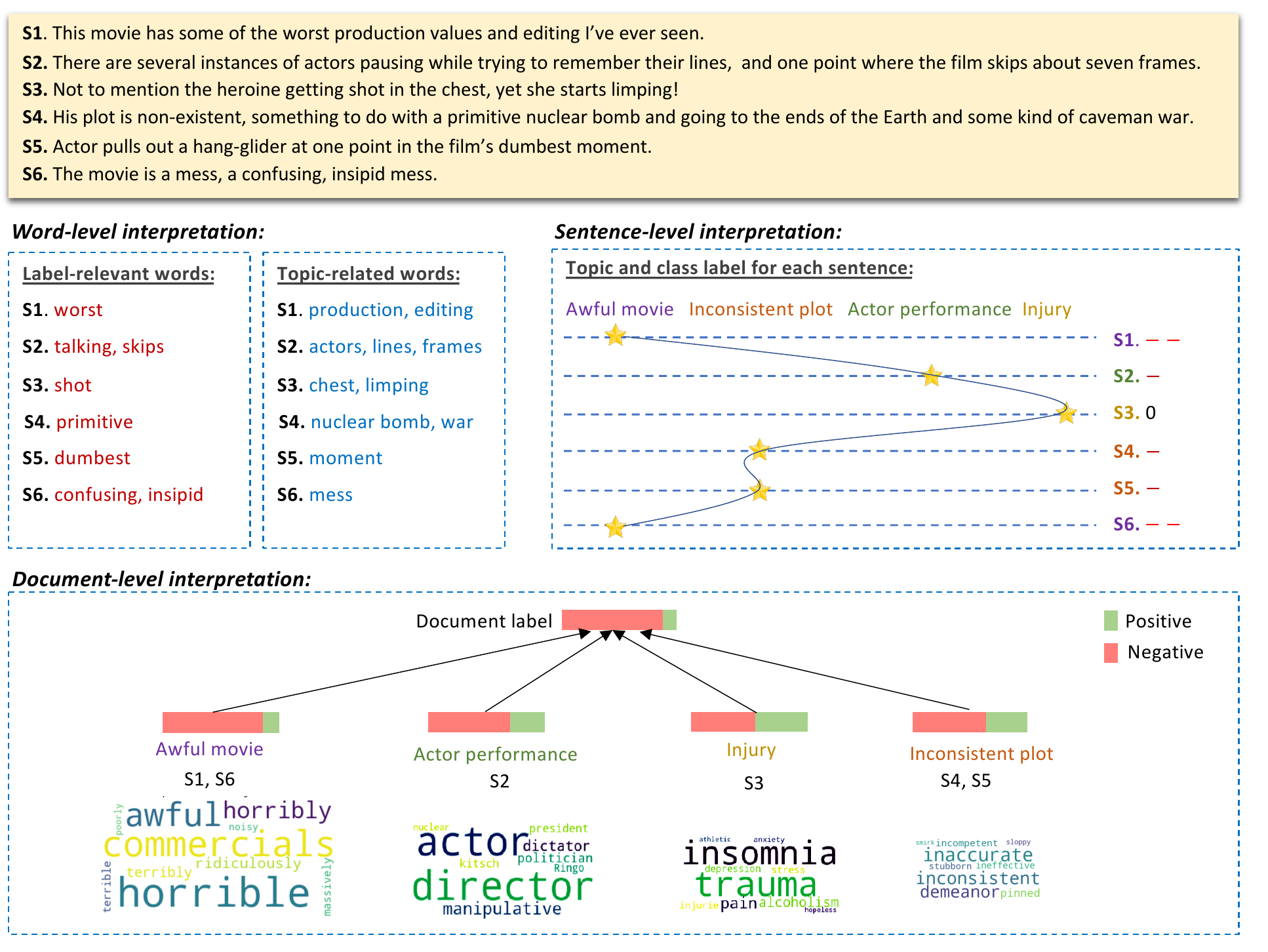}
    \caption{ A \YH{hierarchical} explanation example generated by our proposed method, \textsc{Hint}, on an IMDB review. \textbf{Upper}: a review document. \textbf{Middle}: the left part shows the word-level interpretation including the label-relevant and the topic-related words, while the right part shows the sentence-level interpretation where the topic and class label for each sentence is presented, e.g., Sentence 4 is about the topic `\emph{Inconsistent plot}' with a \emph{negative} polarity. \textbf{Lower}: The document-level interpretation shows the topic partition for the 6 sentences in the review. The size of a word cloud indicates its relative importance in the document. \textcolor{black}{The red and green colour bar above each word cloud indicates the sentiment of the corresponding topic. It is clear that the negative polarity of the review is mainly due to the general negative comments of \emph{awful movie} and more concretely the complaint about the the \emph{actors' performance} and the \emph{inconsistent plot}.} The word-, sentence- and document-level interpretations are generated automatically by our model.}
    \label{fig:imdbex}
\end{figure}

\hqMay{We speculate that a good interpretation model for text classification should be able to identify the key latent semantic factors and their interactions which contribute to the model's final decision. This is often beyond what word-level interpretations could capture. To this end, considering the hierarchical structure of input documents, we propose a novel \YH{Hierarchical} Interpretable Neural Text classifier, called \textsc{Hint}, which can generate interpretations in a hierarchical manner. One example output generated by \textsc{Hint} is shown in Figure~\ref{fig:imdbex} in which a review document consisting of 6 sentences (shown in the upper box) is fed to a classifier for the prediction of a sentiment label. 
Traditional interpretation methods can only identify words which are indicative of sentiment categories as shown in the middle left box in Figure~\ref{fig:imdbex}. However, it is still unclear how these words contribute to the document-level sentiment label, especially when there are words with mixed polarities. Moreover, humans may also be interested in topics discussed in the document and their associated sentiments, and how they are combined to reach the final document-level class label.  
The middle centre box highlights the important topic words in each sentence. Note that traditional post-hoc word-level interpretation methods would not be able to identify these words since they are less relevant to sentiment class labels when considered in isolation. The middle right box in Figure \ref{fig:imdbex} shows the associated topic for each sentence and its respective sentiment label. For example, the sentence $S2$ is associated with the topic `\emph{Actor performance}' and has a \emph{negative} polarity. The lower box shows the topic partition of sentences based on their topic semantic similarities. Such hierarchical explanations (from word-level label-dependent and label-independent interpretations, to sentence-level topics with their associated labels, and finally to the document-level topic partition) are generated automatically from our proposed approach. The only supervision information required for model learning is documents paired with their class labels. }

\yh{As will be shown in the experiments section, our proposed approach achieves comparable classification performance compared to the existing state-of-the-art neural text classifiers when evaluated on 
three document classification datasets. Moreover, it generates interpretations more faithful to model predictions and better understood by humans compared to word-level interpretation methods.} In summary, our contributions are three-fold:
\begin{itemize}[noitemsep,topsep=0pt]
    \item We  propose a neural text classifier with built-in interpretability which can generate hierarchical explanations by identifying both label-dependent and topic-related words at the word-level, detecting topics and their associated labels at the sentence-level, and finally producing the document-level topic and sentiment composition. 
    \item \yh{The evaluation of explanations generated by our approach show that it generates interpretations better understood by humans and more faithfully for model predictions compared to existing word-level interpretation methods.} 
    \item Experimental results show that our proposed approach performs on par with the existing state-of-the-art methods on the three document classification datasets.
\end{itemize}

\section{Related Work}
\label{related work}

\yh{Our work is related to the following lines of research:}

\paragraph{Post-hoc Interpretation}

Post-hoc interpretation methods typically aim to identify the contribution of input attributes or features to model predictions. For example, \citet{DBLP:conf/acl/WuCKL20} proposed a perturbation-based method to interpret pre-trained language models \yh{used for} 
dependency parsing. \citet{DBLP:conf/acl/NiuMDA20} evaluated the robustness and interpretability of neural-network-based machine translation models by 
\yh{slightly} perturbing input. \citet{DBLP:conf/acl/AbdouRBBES20} proposed a new dataset for evaluating model interpretability by seven \yh{different ways of} linguistic perturbations. \citet{DBLP:conf/emnlp/KimYKY20} \yh{argued that interpretation methods that measure the changes in prediction probabilities by erasing word tokens from input text may face the out-of-distribution (OOD) problem. They proposed to marginalise out a token in an input sentence to mitigate the OOD problem. \citet{DBLP:conf/iclr/JinWDXR20} and \citet{DBLP:conf/acl/ChenZJ20} built hierarchical explanations through detecting feature interactions. But their models can only identify sub-text-spans which are important for model decisions and largely focus on sentence-level classification tasks.} 

\paragraph{Self-Explanatory Models}

Different from post-hoc interpretation, \yh{self-explanatory}  
methods aim to generate explanations during model training with \yh{the interpretability naturally}  
built-in.  
Existing work utilises mutual information \cite{DBLP:conf/icml/ChenSWJ18,DBLP:conf/icml/GuanWZCH019}, attention signals \cite{DBLP:conf/acl/ZhouZY20}, Bayesian network \cite{DBLP:conf/acl/ChenDYLH20,DBLP:conf/acl/TangHS20}, or information bottleneck \cite{DBLP:conf/nips/Alvarez-MelisJ18,DBLP:conf/aaai/BangXL0X21} to identify the key attributes or features from input data. 
For example, \citet{DBLP:conf/acl/ZhouZY20} used a variational autoencoder based classifier to identify operational risk in model training. \citet{DBLP:conf/acl/ChenDYLH20} recognised name entities from clinical records and used a Bayesian network to obtain interpretable predictions. \citet{DBLP:conf/acl/ZhangSFL20} proposed an interpretable relation recognition approach by  Bayesian Structure Learning. \citet{DBLP:conf/acl/TangHS20} proposed a rule-based decoder to generate rules for model explanation. \citet{DBLP:conf/emnlp/ZanzottoSROTF20} proposed a kernel-based encoder for interpretable embedding metric to visualise how syntax is used in inference. \citet{DBLP:conf/emnlp/JiangZCST20} \yh{incorporated} regular expressions into recurrent neural network training for cold-start scenarios \yh{in order to} obtain interpretable outputs. 
\yh{\citet{DBLP:conf/emnlp/ChenJ20} proposed variational word masks (VMASK) which are inserted into a neural text classifier after the word embedding layer in order to filter out noisy word-level features, forcing the classifier to focus on important features to make predictions.}

In general, existing \yh{self-explanatory methods} 
mainly focus on tracking the influence of input features \yh{on model outputs and use it as constraints for model learning}. But they ignore the \yh{subtle interplay of input attributes}. In this paper, we propose a novel hierarchical interpretation model, which \yh{can generate interpretations at different granularity levels and achieve classification performance on par with the existing state-of-the-art neural classifiers.} 

\paragraph{Interpretation based on Attentions}

\yh{The attention mechanisms have been widely used in neural architectures applied to various NLP tasks. 
It is common to use attention weights to interpret models' predictive decisions \cite{DBLP:journals/corr/LiMJ16a,lai2019human,de2019bias}. In recent years, however, there have been work showing that attention is not a valid explanation. For example, \citet{DBLP:conf/naacl/JainW19} found that it is possible to identify alternative attention weights after the model is trained, which produced the same predictions. \citet{DBLP:conf/acl/SerranoS19} modified attention weights in already-trained text classification models and analysed the resulting differences in their predictions. They 
concluded that attention cannot be used as a valid indicator for model predictions. While the aforementioned work modified attention weights in a post-hoc manner after a model was trained, 
\citet{DBLP:conf/acl/PruthiGDNL20} proposed to modify attention weights during model learning and produced models whose \emph{actual} weights could lead to deceived interpretations. \citet{DBLP:conf/emnlp/WiegreffeP19} argued the validity of the claim in prior work \cite{DBLP:conf/naacl/JainW19} 
and proposed alternative experimental design to test when/whether attention can be used as explanation. Their results showed that prior work does not disprove the usefulness of attention mechanisms for explainability. }

\section{Hierarchical Interpretable Neural Text Classifier (\textsc{Hint}) }
\label{sec:method}

\begin{figure}[!t]
\centering
\includegraphics[width=\linewidth,trim={0 10 10 10},clip]{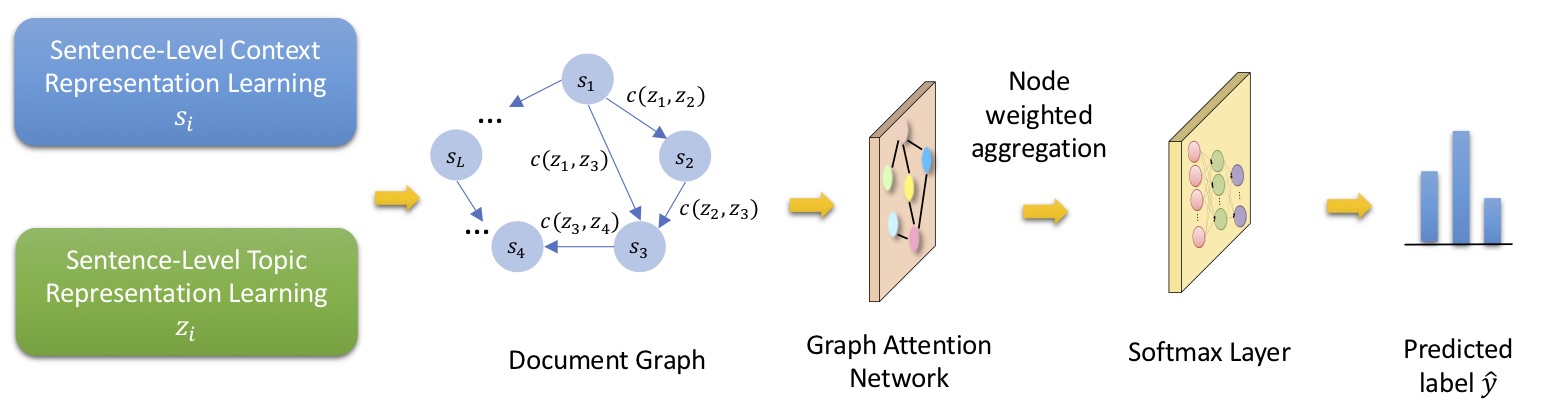}
\caption{ Overall architecture of our proposed dual representation learning framework which consists of two main modules: (a) for a given sentence $i$, the \emph{sentence-level context representation learning} module generates the context representation $\bm{s}_i$ while the \emph{sentence-level topic representation learning} module produces the topic representation $\bm{z}_i$; (b) A document graph is constructed in which each node represents a sentence and its embedding is initialised by its associated context representation $s_i$, the weight of the edge connecting two nodes is determined by the similarity between their corresponding topic representations, denoted as $c(\bm{z}_i, \bm{z}_j)$. Node representations are updated by a Graph Attention Network and the document representation is derived by \YH{weighted aggregation of} the node representations. Finally, the document representation is fed to a softmax layer to predict the class label $\hat{y}$.}
\label{fig:modelOverview}
\end{figure}

Our proposed \textbf{H}ierarchical \textbf{I}nterpretable \textbf{N}eural \textbf{T}ext (\textsc{\YH{Hint}}) classification model is shown in Figure~\ref{fig:modelOverview}. For each sentence in an input document, a dual representation learning module (\textsection\ref{sec:dual_module}) is used to generate the 
contextual representation \yh{guided by the document class label} and the latent topic representation, \yh{from which the word-level interpretations can be generated.} 
\hqDec{To aggregate the sentences with similar topic representations, we create a fully-connected graph (\textsection\ref{sec:doc_graph}), whose nodes are initialised by the sentence contextual representations and edge \hqMay{weights} are topic similarity values of the respective sentence nodes. Sentence interactions are captured by a single-layer Graph Attention Network to derive the document representation for classification.} 
In what follows, we describe each of the modules of \textsc{\YH{Hint}} in detail. The notations used in this article are shown in Table~\ref{tab:notations}.

\begin{table}[!t]
\caption{ Notations used in the article.}
\small
\begin{tabular}{lp{10.5cm}}     \toprule
\textbf{Symbol} & \textbf{Description} \\ \midrule
\multicolumn{2}{l}{Sentence Representation Learning Module} \\ \midrule
$N$         & The dimension of input word embeddings and learned sentence embeddings. \\
$x_{ij} \in \mathbb{R}^{N}$         & The input embedding of $j$-th word in $i$-th sentence.         \\ 
$\bm{s}_{i} \in \mathbb{R}^{N}$         & The learned  embedding of $i$-th sentence.        \\ 
$\mbox{biLSTM}_{\phi}$         & The bidirectional LSTM encoder with learnable parameter set $\phi$.      \\
$u_{ij} \in \mathbb{R}^{N/2}$         & The attention vector for $j$-th word in $i$-th sentence.      \\
$a_{ij} \in \mathbb{R}$  & The attention signal for $j$-th word in $i$-th sentence. 
\\ \midrule
\multicolumn{2}{l}{Topic Representation Learning Module} \\ \midrule
$K$         & The dimension of topic embeddings. \\
$\beta_{ij}$         & The topic based attention signal for $j$-th word in $i$-th sentence.          \\ 
$\bm{\omega}$ & The parameter set in topic based attention learning, which includes $W_{\mu},W_{\omega}\in \mathbb{R}^{N \times K}$ and $b_{\mu},b_{\omega}\in \mathbb{R}^{K}$.         \\ 
$\bm{r}_i\in \mathbb{R}^{N}$         & The sentence representation of the $i$-th sentence derived based on the word-level topic attentions $\beta_{ij}$.      \\
$\bm{z}_i\in \mathbb{R}^{K}$         & The topic representation of the $i$-th sentence.      \\
$\bm{r}'_i\in \mathbb{R}^{N}$         & The reconstructed sentence representation of the $i$ sentence based on the learned autoencoder.      \\  
$\mathcal{R}_i$, $\lambda_i$ & The regularisation term and its corresponding weight. \\
$z_{ik}$ & The probability that a sentence $i$ belongs to the $k$-th topic, also represented as $P(t_k | \bm{s}_i)$.\\
$g_k^d$ & The occurrence probability of the $k$-th topic in document $d$, also defined as $P(t_k | d)$.\\
\midrule
\multicolumn{2}{l}{Document Representation Learning Module} \\ \midrule
$c_{ij}$         & The  similarity between the $i$-th and $j$-th sentence based on the learned topic representation.      \\ $e_{ij}$         & The  static edge weight derived by normalising $c_{ij}$. \\
$\bm{s}^{l}_{i}$         & The representation of the $i$-th sentence learned by the graph attention network in the $l$-th iteration, where $\bm{s}^{0}_{i}$ is initialised by $\bm{s}_{i}$. \\ 
$\bm{w}_d$ & The representation of document $d$. \\
$\eta_a, \eta_b$         & The weight of different term in the loss function for document representation learning.  \\  \bottomrule[1pt]
\end{tabular}
\label{tab:notations}
\end{table}

\subsection{Dual Module for Sentence-Level Representation Learning}
\label{sec:dual_module}

The dual module captures the sentence contextual and latent topic information separately. In particular, we hope that the sentence-level context representation would capture the label-dependent semantic information, while the sentence-level topic representation would encode label-independent semantic information shared across documents regardless of their class labels. 

\subsubsection{Context Representation Learning}
\label{sec:contextualLearning}

\begin{figure}[!t]
\centering
\begin{subfigure}{.5\textwidth}
  \centering
  \includegraphics[width=0.6\linewidth]{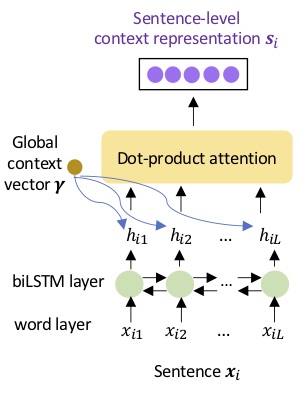}
  \caption{Context representation learning.}
  \label{fig:sentCtxRepLearning}
\end{subfigure}
\begin{subfigure}{.5\textwidth}
  \centering
  \includegraphics[width=\linewidth]{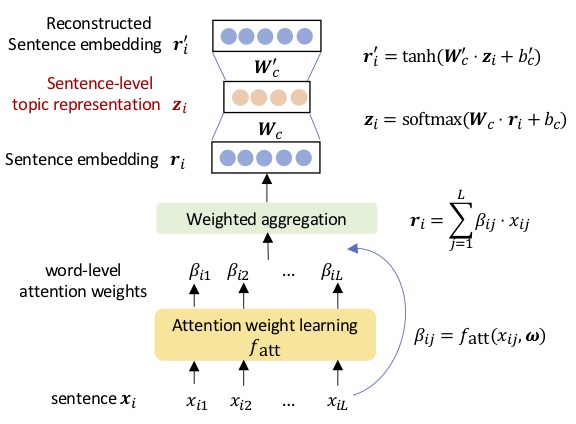}
  \caption{Topic representation learning.}
  \label{fig:sentTopicRepLearning}
\end{subfigure}
\caption{ Sentence-level context and topic representation learning.}
\end{figure}

In the \emph{sentence-level context representation learning module} shown in Figure \ref{fig:sentCtxRepLearning}, the goal is to capture the contextual representation of a sentence with word-level label-relevant features. We choose a bidirectional LSTM (biLSTM) network, which captures the contextual semantics information conveyed in a sentence, with an attention mechanism which can capture the task relevant weights for interpretation \cite{DBLP:conf/naacl/YangYDHSH16}.

Assuming that a document $d$ contains $M_d$ sentences, $\bm{w}_d=\{s_1,s_2,...s_{M_d}\}$, 
and each sentence indexed by $i$ contains $L$ words with each word represented by a pre-trained $N$-dimensional word embedding, $\bm{x}_i=\{x_{i1},x_{i2},...x_{iL}\}, x_{ij}\in\mathbb{R}^N$, then the hidden representation for each word $x_{ij}$, denoted as $h_{ij}\in\mathbb{R}^N$, is obtained by: 
\begin{equation}
    \{h_{i1},h_{i2},...h_{iL}\} =\mbox{biLSTM}_{\phi}(x_{i1},x_{i2},...x_{iL}),
\end{equation}
\noindent where $h_{ij}$ is the hidden representation of input word $x_{ij}$ learned by the encoder $\mbox{biLSTM}_{\phi}$ with learnable parameters $\phi$. Based on the learned word representation, we aggregate the context representation of sentence $i$, denoted as $\bm{s}_i$, by a two-layer Multi-Layer Perceptron (MLP) based attention:
\begin{equation}
\begin{gathered}
u_{ij} = {\rm tanh}(W_s \cdot h_{ij} + b_s),\quad
\alpha_{ij} = \frac{{\exp}(\gamma^\intercal \cdot u_{ij})}{\sum_{j'} {\exp} (\gamma^\intercal \cdot u_{ij'})},\quad
\bm{s}_i = \sum^{L}_{j=1} \alpha_{ij} \cdot h_{ij},
\label{eq:att}
\end{gathered}
\end{equation}
\noindent where  
$W_s\in \mathbb{R}^{N \times N/2}, b_s \in \mathbb{R}^{N/2}$ are learnable parameters for the first layer MLP with the activation function  $\mathtt{tanh}$. The output of the first layer MLP, $u_{ij} \in \mathbb{R}^{N/2}$, is the attention vector of the $j$-th word in the $i$-th sentence. In the second layer MLP, we use an inner product based mapping function with $\mathtt{softmax}$ normalisation to capture the attention signal of $\alpha_{ij}$ for $u_{ij}$. Here, 
$\gamma$ is a learnable vector in the second-layer MLP and is shared among all sentences, which can be considered as a centre point for label relevant representation in the latent space, or a global context vector. The similarity between $u_{ij}$ (i.e., the word representation after the first layer MLP) and $\gamma$ reflects the importance of the corresponding word in the classification. 
We use $\bm{s}_i$ to denote the learned context representation for the $i$-th sentence in a document $d$.

\yh{The aforementioned approach in producing the sentence-level contextual representations is a typical way in encoding sentence semantics. When used in building neural classifiers, we would expect such representations implicitly capture the class label information. More concretely, its word-level attention weights can be used to identify words which are important for text classification decisions. Taking sentiment classification as an example, as have been previously shown in hierarchically-stacked LSTM or GRU networks, words with higher attention weights are often indicative of polarities \cite{DBLP:conf/naacl/YangYDHSH16} .} 

\subsubsection{Topic Representation Learning}
\label{sec:topic_learning}

Sentence-level contextual representations learned in Section \ref{sec:contextualLearning} implicitly inject label information to $\bm{s}_{i}$ and capture label-dependent word features. Here, we propose to use a label-independent approach to capture the hidden relationships between input words to infer latent topics, which could be subsequently used to determine their potential contributions to class labels, in order to enhance the generalisation and interpretability.

We propose to use a Bayesian inference based autoencoder to learn the sentence-level topic representation, as shown in Figure~\ref{fig:sentTopicRepLearning}. More concretely, we assume the conditional probability of $k$-th topic $t_k$ given a sentence $\bm{x}_{i}$, denote as $P(t_k|\bm{x}_{i})$, is obtained by the conditional probability of the corresponding topic given its constituent words $x_{ij}$ by:
\begin{equation}
    P(t_k|\bm{x}_{i}) = \sum_{j=1}^{L} P(t_k|x_{ij}) \cdot P(x_{ij}|\bm{x}_i).
\vspace{-0.4cm}
\end{equation}

In this equation, $P(t_k|x_{ij})$ can be learned by an autoencoder and $P(x_{ij}|\bm{x}_i)$ is obtained by a Bayesian inference approach. In particular, we denote $P(x_{ij}|\bm{x}_i)$ as $\beta_{ij}$, which can be considered as the topic-based attention weight of a word $x_{ij}$ in a sentence $\bm{x}_i$. It is a latent variable and its value is given by a stochastic generative process:
\begin{equation}
\label{eq:beta}
    \beta_{ij} = f_{\mbox{att}}(x_{ij}, \bm{\omega}), \quad \bm{\omega} \sim \mathcal{N}(\mu_{\omega}, \sigma_{\omega}^2),
\end{equation}
\noindent where $x_{ij}$ is the embedding of the $j$-th word in $i$-th sentence, $\bm{\omega}$ denotes the parameters for attention weight learning.  
More generally, we aim to learn $p(\bm{\omega} | \mathcal{D}) \propto p(\mathcal{D} | \bm{\omega})p(\bm{\omega})$ where $\mathcal{D}$ denotes all the training documents, and $\bm{\omega}$ is sampled from the variational posterior $q(\bm{\omega} | \mathcal{D})$ which also assumes following a Gaussian distribution and can be approximated by a neural network, i.e., 
\begin{align}
\bm{\mu}_{\omega} &= f_{\mu}(x_{ij}) = \mbox{sigmoid}(W_{\mu}\cdot x_{ij} + b_{\mu}),\\
\bm{\sigma}_{\omega}^2 &= f_{\sigma}(x_{ij}) = \mbox{sigmoid}(W_{\sigma}\cdot x_{ij} + b_{\sigma}),\\
\epsilon  \sim &\mathcal{N}(0,\mathbf{I}),\quad
f_{ \mbox{att} }(x_{ij},\bm{\omega}) = \bm{\mu}_{\omega} + \bm{\sigma}_{\omega} \cdot \epsilon.
\end{align}\label{eq:sample}
\noindent Here, $f_{\mu}$, $f_{\sigma}$ are MLP-based approximation to obtain the mean and variance of the input word representation, $W_{\mu},W_{\omega}\in \mathbb{R}^{N \times K}$ and $b_{\mu},b_{\omega}\in \mathbb{R}^{K}$ are learnable parameters in the MLP layer, $\mathbf{I}$ is the identity matrix. The parameters in these layers are shared across all input words. 

\yh{Once the word-level attention weights are learned, the sentence embedding, denoted as $\bm{r}_i\in \mathbb{R}^{N}$ is obtained by $\bm{r}_i = \sum^L_{j=1} \beta_{ij} \cdot x_{ij}$. We then feed the sentence-level representation $\bm{r}_i$ into an autoencoder to generate the reconstructed representation $\bm{r}'_i$: }
\begin{equation}
\bm{z}_i = \mbox{softmax}(\bm{W}_c \cdot \bm{r_i} + \bm{b}_c),\quad
\bm{r'}_i = \mbox{tanh}(\bm{W'}_c \cdot \bm{z_i} + \bm{b'}_c),
\label{eq:latent_topic}
\end{equation}
\noindent where $\bm{W}_c$,$\bm{W'}_c\in \mathbb{R}^{N\times K}$, $\bm{b}_c$, 
$\bm{b}'_c \in\mathbb{R}^{K}$ are learnable parameters which can be used to generate topics,  
$\bm{z}_{i}\in \mathbb{R}^{K}$ is the hidden topic vector which is considered as the sentence-level topic distribution for $i$-th sentence and $\bm{r}'_{i} \in \mathbb{R}^{N}$ is the reconstructed sentence representation for $i$-th sentence based on the corresponding topic representation $\bm{z}_i$.
Since $p(\mathcal{D} | \bm{\omega})$ is intractable, we resort to neural variational inference to maximise the Evidence Lower BOund (ELBO) that
\vspace*{-0.5cm}
\begin{equation}
    \mathcal{L}_{e}(\bm{w}_d) = \sum_{i=1}^{M_d} \log p(\bm{r}'_i |x_{ij}, \bm{\omega}) 
   -  D_{KL}(q(\bm{\omega} | \mathcal{D}) ||  p(\bm{\omega})),
    \label{eq:attKL}
\end{equation} 
\noindent where $\bm{w}_d$ denotes the representation for document $d$. 
The first term denotes the reconstructed sentence representation, and the second term is the KL diversity measuring the difference between the variational posterior and the prior distribution. The posterior distribution is learned from the training corpus $\mathcal{D}$, while the prior distribution of $\bm{\omega}$ is a normal distribution\footnote{ Note that $\bm{\omega}$ is shared among all inputs, which is different from typical latent variable models in which a local latent variable is associated with each individual input.}. 

 \paragraph{Regularisation Terms}

In the following, we introduce a number of regularisation terms used in our model.

\noindent{\underline{Orthogonal regularisation}}. 
To make learned sentence-level topic representation different from the sentence-level context representation, we simultaneously minimise the inner product between the reconstructed sentence representation $\bm{r}_i'$ and the context representation $\bm{s}_i$. 
Hence, we define an orthogonal regularisation term below:
\begin{equation}
\mathcal{R}_1(\bm{w}_d) =  \sum_i^{M_d} {||  \bm{r}_i' \cdot \bm{s}_i ||}_2 \label{eq:topicOrthogonal}
\end{equation}

\noindent{\underline{Topic uniqueness regularisation}}. 
Note that the learned topics \yh{might be redundant, i.e., different topics might contain many overlapping words}. To ensure the diversity of the resulting topics learned, we add a regularisation term to the objective function to encourage the uniqueness of each topic embedding.
\begin{equation}
\mathcal{R}_2(\bm{w}_d) =  {|| \bm{W'}_c \cdot \bm{W'}^\intercal_c - \mathbf{I} ||}_2
\label{eq:topicUniqueness}
\end{equation}
\noindent where $\mathbf{I}$ is the identity matrix, and $\bm{W'}_c$ is the  decoder matrix defined in Eq. (\ref{eq:latent_topic}), where each column can be extracted as the representation of a topic. $\mathcal{R}_2(\bm{w}_d)$ reaches its minimum value when the dot product between any two different topic representations is zero. 

\noindent{\underline{Topic discrepancy regularisation}}.
Based on the topic representations, $\bm{z}_{i}$, we can essentially partition text into different groups. Inspired by \citet{DBLP:conf/icml/JohanssonSS16}, we propose another regularisation term to re-weigh different partitions. 

Intuitively, 
we want to reduce the discrepancy between different latent topics weighted by the posterior probability of topics given text in order to prevent the learner from using ``unreliable'' topics of the data when trying to generalise from the factual to the counterfactual domains. For example, if in our movie reviews, very few people mentioned the \YH{topic} of `\emph{source effect}', inferring the attitude towards this \YH{topic} is highly prone to error. As such, the importance of this \YH{topic} should be down weighted.

Without loss of generality, assuming an input sentence $\bm{x}_i$ contains $L$ words,  $\bm{x}_i =\{x_{i1},x_{i2},...,x_{iL}\}$, where $x_{ij}$ is the word embedding of the $j$-th word in $i$-th sentence. Each sentence is mapped to a latent topic distribution with $K$ dimensions, $\bm{z}_i=\{z_{i1},z_{i2},...,z_{iK}\}$, with each of its elements representing the probability that the input sentence $i$ belongs to the $k$-th topic, $z_{ik}=P(t_k|\bm{x}_i)$. 
The occurrence probability of the $k$-th topic in document $d$ is defined as:
\begin{equation}
    P(t_k|d) = \frac{1}{M_d}\sum_{i=1}^{M_d}P(t_k|\bm{x}_i)  
    =\frac{1}{M_d}\sum_{i=1}^{M_d}z_{ik} 
    = g_{k}^d,
\label{eq:occurrence}
\vspace*{-0.2cm}
\end{equation}



\noindent Inspired by \citet{DBLP:conf/icml/JohanssonSS16}, in which the discrepancy is defined as a function of the distance between the weighted population means, 
we define the discrepancy between two topics, $t_a, t_b$, in a document $d$ as the distance between two topic representations weighted by their occurrence probabilities in document $d$:
\begin{align}
    disc (t_a, t_b ) &= 1 - \mbox{cos}\big(P(t_a|d)\zeta_{t_a} , P(t_b|d)\zeta_{t_b}\big) \nonumber \\
    &\varpropto 1 - ( g_a^d \cdot W_{ca} ) (g_b^d \cdot W'_{cb})^\intercal,
    \label{eq:disc-term}
\vspace*{-0.2cm}
\end{align}

\noindent where \mbox{cos}$(\cdot)$ denotes the cosine similarity function, $\zeta_{t_a}$ denotes the representation of the topic $t_a$ (similarly for $\zeta_{t_b}$), and is equivalent to  $W_{ca}$, the $a$-th column of the encoder matrix. A brief explanation of why $W_{ca}$ can be considered as the representation of topic $t_a$ will be given in Section \ref{sec:generate_interpretation}. In Eq. (\ref{eq:disc-term}), the topic representations $W_{ca}$ and $W_{cb}$ are global and are shared across all documents, while the topic occurrence probabilities $g_a^d$ and $g_b^d$ are local and are specific to document $d$. 
To understand the effect of applying a regularisation term defined in Eq. (\ref{eq:disc-term}), we illustrate below the derivation of the gradient on topic $t_a$'s representation $\zeta_{t_a}$. First, assuming the regularisation term defined based on Eq. (\ref{eq:disc-term}) is $\mathcal{R}(disc (t_a, t_b ))$, then the corresponding gradient on topic $t_a$'s representation $\zeta_{t_a}$ is $\frac{\partial \mathcal{R}\big(disc (t_a, t_b )\big)}{\partial disc (t_a, t_b )} \cdot \frac{\partial disc (t_a, t_b )}{\partial \zeta_{t_a}}$. Without loss of generality, we do not give a specific formula of the regularisation function here and only focus on the second component, $\frac{\partial disc (t_a, t_b )}{\partial \zeta_{t_a}}$: 


\begin{align}
    \frac{\partial disc (t_a, t_b )}{\partial \zeta_{t_a}} &= - \big( \frac{\partial g_a^d \cdot W_{ca}  }{\partial W_{ca}} \big) \cdot (g_b^d \cdot W_{cb})^\intercal  \nonumber \\
    &= - \big( \frac{\partial \sum_{i=1}^{M_d}  \bm{s}_i \cdot  W_{ca}^\intercal \cdot W_{ca} }{\partial W_{ca}} \big) \cdot \frac{(g_b^d \cdot W_{cb})^\intercal}{M_d} \nonumber \\
    &= -2 \cdot g_b^d  \frac{\sum_{i=1}^{M_d} \bm{s}_i \cdot W_{ca}  \cdot W_{cb}^\intercal}{M_d} \nonumber \\
    &\varpropto -2 \cdot g_b^d \cdot \mbox{cos}(\zeta_{t_a}, \zeta_{t_b}) \frac{\sum_{i=1}^{M_d} \bm{s}_i }{M_d},
\vspace*{-0.2cm}
\end{align}

The above result state that if two input topics have similar representations, or have higher occurrence probabilities in the current document, they will obtain larger updates which push their representations closer to the representation of the current document calculated based on the mean pooling of its constituent sentences. Essentially, the discrepancy term separates the document representations based on the topic similarities, then updates the corresponding topic distribution based on the topic occurrence probabilities and the document representation.

We can extend the discrepancy term to cater for all possible topic pairs as:
\begin{gather}
Disc_d = 
\left[
\begin{matrix}
 disc (t_1, t_1 )    &  disc (t_1, t_2 )      & \cdots &  disc (t_1, t_K )     \\
 disc (t_2, t_1 )    &  disc (t_2, t_2 )      & \cdots &  disc (t_2, t_K )     \\
 \vdots & \vdots & \ddots & \vdots \\
  disc (t_K, t_1 )    &  disc (t_K, t_2 )      & \cdots &  disc (t_K, t_K )      \\
\end{matrix}
\right]=  {\textbf 1}_{K \times K} - P_m(d) \odot  \bm{W}'_c {\bm{W}'_c}^\intercal ,
\label{eq:disc-joint}
\end{gather}
\noindent where ${\textbf 1}_{K \times K}$ is a $k \times k$ matrix in which all elements are 1, $\odot$ is the element-wise product, $P_m(d)$ is defined as:
\begin{gather}
 P_m(d) = 
\left[
\begin{matrix}
 g_1^d \cdot  g_1^d     &  g_1^d \cdot  g_2^d      & \cdots &  g_1^d \cdot  g_K^d     \\
 g_2^d \cdot  g_1^d      &  g_2^d \cdot  g_2^d     & \cdots &  g_2^d \cdot  g_K^d      \\
 \vdots & \vdots & \ddots & \vdots \\
 g_K^d \cdot  g_1^d     &  g_K^d \cdot  g_2^d     & \cdots &  g_K^d \cdot  g_K^d     \\
\end{matrix}
\right],
\label{eq:topicJoint}
\vspace*{-0.2cm}
\end{gather}

\noindent We can then define the discrepancy term based regularisation by $l_2$ norm as $\lVert {\textbf 1}_{K \times K} - P_m(d) \odot  \bm{W}'_c {\bm{W}'_c}^\intercal \rVert_2$. 
The gradient of such a regularisation term with respect to a specific topic representation would guide its movement towards the centroid of the other topics weighted by their occurrence probabilities defined in Eq. (\ref{eq:occurrence}). While the topic uniqueness regularisation term defined in Eq. (\ref{eq:topicUniqueness}) aims to ensure the orthogonality among topics, the discrepancy term based regularisation will push the representations of major topics closer to the input document representation. 




In practice, to achieve the balance between two regularisation terms, we combine the discrepancy term based regularisation 
with the topic uniqueness regularisation defined in Eq. (\ref{eq:topicUniqueness}) as:
\begin{equation}
    \mathcal{R}_2(\bm{w}_d) = \lVert  \bm{W}'_c \cdot{\bm{W}'_c}^\intercal - \big(\alpha  \cdot  \mathbf{I} + (1 - \alpha ) \cdot P^{-1}_m(d)\big)  \rVert_2, 
    \label{eq:disc}
\vspace*{-0.2cm}
\end{equation}
\noindent where $\alpha \in (0,1)$ determines the contribution of the topic uniqueness term. For any $P^{-1}_m(d)_{ij}$, which denotes the the $i$-th row and $j$-th column element in $P^{-1}_m(d)$, $P^{-1}_m(d)_{ij}=1/(g^d_i \cdot g^d_j)$. In our experiments, we set $\alpha = 0.5$.

\noindent{\underline{Final objective function}}.
The final objective function $L_{\mbox{topic}}$ for sentence-level topic representation learning is defined below, where $\lambda_{1}$ and $\lambda_{2}$ are used to control the relative contributions of different terms.
\begin{equation}\label{eq:topicObj}
    \mathcal{L}_{\mbox{topic}}(\bm{w}_d) = \mathcal{L}_e(\bm{w}_d)+ \lambda_{1} \cdot\mathcal{R}_1(\bm{w}_d) + \lambda_{2} \cdot\mathcal{R}_2(\bm{w}_d)
\vspace*{-0.2cm}
\end{equation}

\subsection{Document Modelling}
\label{sec:doc_graph}
After obtaining the sentence-level context representations $\bm{s}_{i}$ and latent topic representations $\bm{z}_{i}$, \yh{the next step is to aggregate such representations to derive the document-level representation for classification.} 
\paragraph{Graph Node Update}
\yh{We represent each document by a graph in which} the nodes represent sentences $\{\bm{s}_{i}\}_{i=1}^{M_{d}}$ in the document $d$ and the edges linking every two nodes measure their topic similarities. Each graph node is initialised by its respective sentence-level context representation $\bm{s}_{i}$. The topic similarity $c_{ij}$ between the $i$-th sentence and $j$-th sentence is defined as the inner-product of their latent topic vectors, $ c_{ij} = \bm{z}_{i}^\intercal \bm{z}_{j}$. 

Graph Attention Networks~(GATs) is applied to update the graph nodes. \hq{Classic GATs learn attention weights via applying self-attention on node features, and update the edge weights during training. Here, we leverage the normalised topic similarity $e_{ij}=\mathtt{softmax}(c_{ij})$ as the static edge weight and aggregate the sentences sharing the similar topics. In this way, \hqMay{sentences linked with larger weight edges are topically more similar}. The graph node features are updated as follow:}
\begin{equation}\label{eq:graph_update}
    \bm{s}_{i}^{\ell+1} = \mathtt{Relu}(\sum_{j\in \mathcal{N}_{i}}e_{ij}\bm{W}\bm{s}_{j}^{\ell}), 
\vspace*{-0.3cm}
\end{equation}
\noindent \yh{where $\bm{s}_{i}^{\ell+1}$ denotes the hidden representation of node (or sentence) $\bm{x}_i$ in the $(\ell+1)$th iteration and $\bm{s}_{i}^{0}$ is initialised by the context sentence representation $\bm{s}_i$ learned from Context Representation Learning module, $\mathcal{N}_{i}$ denotes the neighbours of node $i$}, $\bm{W}$ is the learnable weight matrix for the graph nodes.

\paragraph{Document Classification}
For a document $d$ given the last layer output from its graph, $\{\bm{s}_{i}^{L}\}_{i=1}^{M_d}$, we average the $M_d$ node representations as the document representation, $\bm{w}_d = (\bm{s}_{1}^{L}+\bm{s}_{2}^{L}...+\bm{s}_{M_d}^{L})/M_d$. The document representation is fed to our classification layer (i.e., softmax) to generated the predicted outputs, $\hat{y}=\mathtt{softmax}(\bm{w}_d)$. The classification loss is defined as:
\begin{equation}
\mathcal{L}_c(\bm{w}_d) = -\sum_{c=1}^C y_c \cdot \log(\hat{y}_c),
\vspace*{-0.2cm}
\end{equation}
where $C$ denotes the total number of class categories. 
The final loss function is defined as:
\begin{equation}
\mathcal{L}_{final}(\bm{w}_d) =  \eta_a \cdot \mathcal{L}_{\mbox{topic}}(\bm{w}_d) + \eta_b \cdot \mathcal{L}_c(\bm{w}_d),
\vspace*{-0.3cm}
\end{equation}
\noindent where $\eta_a$ and $\eta_b$ are the weights to control the contribution of the respective loss to the final objective function.

\section{Model Interpretation Generation}
\label{sec:generate_interpretation}
\hq{For a given document $d$, the proposed model can not only predict a label, but also generate a \YH{hierarchical} interpretation for its prediction. Taking the document in Figure~\ref{fig:imdbex} as an example, we will elaborate below how to generate the word- and sentence-level explanations, as well as} 
\hqMay{how to aggregate the hierarchical information to generate the final model prediction.}

\paragraph{Word-Level Interpretation Generation}
\label{sec:wordlevel_ex}
\hq{The dual module learns a context representation and a topic representation of a sentence.} \yh{The approach in producing the context representations is a typical way in encoding sentence semantics (\textsection\ref{sec:contextualLearning}). When used in building neural classifiers, we would expect such representations implicitly capture the class label information. More concretely, its word-level attention weights can be used to identify words which are associated with the class label. 
\hqDec{For the example shown in Figure~\ref{fig:imdbex}, label-relevant words such as `\emph{worst}' is indicative of the \emph{Negative} polarity.} The latent topic learning module~(\textsection\ref{sec:topic_learning}) aims to capture latent topics shared across all documents regardless of their class labels. The word-level attention weights are generated by a stochastic process as shown in Eq. (\ref{eq:beta}). 
It can be observed that words identified in this way are more topic-related (such as `\emph{production}' and `\emph{actors}') and are less relevant to the class label. }

\yh{To extract topic words (the word cloud in Figure~\ref{fig:imdbex}), we first multiply the word embeddings matrix $E\in\mathbb{R}^{V\times N}$ with the weight matrix $\bm{W}_c\in\mathbb{R}^{N\times K}$ from the topic encoder network (Eq. (\ref{eq:latent_topic}) in \textsection\ref{sec:topic_learning}), where $V$ denotes the vocabulary size,
$N$ is the word embedding dimension, and $K$ is the topic number~\footnote{ We can also use the decoder weight matrix $\bm{W'}_c$, which is symmetry to the $\bm{W}_c$.}. \hqMay{From the resulting matrix $\bm{\pi}\in\mathbb{R}^{V\times K}$,} we can then extract the top $n$ words from each topic dimension as the topic words~\cite{Chaney_Blei_2021}. In the following, we explain why each column in $\bm{\pi}$ can be considered as a topic.}

In Section \ref{sec:dual_module}, we assume the topic distribution is obtained by an encoder-decoder formulation in Eq. (\ref{eq:latent_topic}), its topic representation for sentence $\bm{x}_i$, denote as $\bm{z}_i$, is a $K$-dimension vector of $\bm{z}_i=\{z_{i1},z_{i2},...,z_{iK}\}$, with each of its elements representing the probability that the input sentence belongs to the $k$-th topic, 
$z_{k}=P(t_k|\bm{x})$,
\noindent and the encoder layer can be rewritten as a softmax function which generates a probability:

\begin{equation}
    z_{ik} = \frac{\exp\big(W_{ck}^\intercal \cdot \sum_j x_{ij} \cdot P(x_{ij}|\bm{x}_i)+b_{ck}\big)}{\sum_m\exp\big(W_{cm}^\intercal \cdot \sum_j w_j \cdot P(x_{ij}|\bm{x}_i)+b_{cm}\big)}, \nonumber
\end{equation}
\noindent where $W_{ck}$ is the $k$-th column of the encoder matrix $\bm{W}_c=\{W_{c1},W_{c2},...,W_{cK}\}$, and $\bm{b}_c=\{b_{c1},b_{c2},...,b_{cK}\}$ is the bias term. We then have:

\begin{equation}
    z_k \varpropto W_{ck}^\intercal \cdot \sum_jx_{ij} \cdot P(x_{ij}|\bm{x}_{i}) +b_{ck} \varpropto W_{ck}^\intercal \cdot \sum_j  x_{ij} \cdot P(x_{ij}|\bm{x}_i)\\
\label{eq:cond}
\end{equation}
\noindent Since the activation function in our network is bijection, we can simply take $z_k \varpropto W_{ck}^\intercal \cdot \bm{x}_i$ to guarantee that Eq. (\ref{eq:cond}) is correct. Therefore, we can use the corresponding column in the encoder matrix to search the whole vocabulary to identify the top-$n$ words in each topic.



\paragraph{Sentence-Level Interpretation Generation}

\yh{From the sentence-level latent topic representation, we can identify the most prominent topic dimension in the hidden topic vector $z_{i}$ and use it as the topic label for each sentence. As has been illustrated in the lower part (word cloud) in Figure~\ref{fig:imdbex}, the topics that correspond to the \hqDec{six sentences can be summarised as `\emph{Awful movie}', `\emph{Actor performance}', `\emph{Injury}', and `\emph{Inconsistent Plot}', from left to right.} \hqMay{
Here, we represent each topic as a word cloud, which contains the top-10 topic-associated words from the corpus vocabulary as shown in Figure \ref{fig:imdbex}. The topic labels are manually assigned for better illustration. We can also automatically generate topic labels by selecting the most relevant phrase from the document to represent each topic (see in Figure~\ref{fig:case_study}). Specifically, for each topic, we first find the most relevant sentence according to the sentence-topic distribution, i.e., $\mathcal{T}^{M\times K} = S^{M\times d}\times (Z^{K\times d})^{\intercal}$, where $M$ is the number of sentences in a document, $d$ is the dimension of a sentence representation, $S$ and $Z$ are the sentence contextual  representations and the topic representations, respectively. Then, we extract the key phrase from the sentence via the Rapid Automatic Keyword Extraction \textsc{RAKE} algorithm\footnote{ \url{https://pypi.org/project/rake-nltk/}}.} We infer the \YH{class} label of each sentence by feeding }
\hq{
the sentence contextual representation into the classification layer of \textsc{\YH{Hint}} and obtain probabilities of \YH{class} labels, thus obtaining its \YH{class-associated} intensity.}

\paragraph{Document-Level Interpretation Generation} \yh{Once the topic and class label for each sentence is obtained, we can aggregate sentences based on the similarity of their latent topic representations. The contextual representation of the document is obtained by taking the weighted aggregation of its constituent sentence contextual representations, \YH{where the weights are the topic similarity values}. As sentences are assigned to various topics, we can easily study how topics and their associated class labels change throughout the document. In addition, we can also infer the most prominent topic in the document.
}

\section{Experimental Setup}
\label{experiment}

\subsection{Datasets}
We 
\yh{conduct experiments on three \YH{English} document datasets,} including two review datasets: patient reviews extracted from \textsl{Yelp}\footnote{ \url{https://www.yelp.com/dataset}} and the \textsl{IMDB} movie reviews~\cite{maas2011learning}, as well as the \textsl{Guardian News} Dataset\footnote{ https://www.kaggle.com/sameedhayat/guardian-news-dataset/tasks}. For Yelp reviews, we retrieve patient reviews based on a set of predefined keywords\footnote{ All the keywords are listed in Appendix B.}. Each review is accompanied with keywords indicating its associated healthcare categories. Since the majority of reviews have ratings of either 1 or 5 stars, we only keep the reviews with 1 and 5 stars as negative and positive instances, respectively. The IMDB dataset also has two class categories (\emph{positive} and \emph{negative}). Yelp has positive reviews twice of negative ones while IMDB has a balanced class distribution. As the IMDB dataset does not \yh{provide the} 
train/test split, we follow the same split proportion as that in the implementation of Scholar\footnote{ https://github.com/dallascard/scholar}. The Guardian News dataset contains 5 categories, i.e., \emph{Sports}, \emph{Politics}, \emph{Business}, \emph{Technology} and \emph{Culture}, from which nearly 40\% documents are in the \emph{Sports} category and less than 10\% and 5\% documents are in the \emph{Technology} and \emph{Culture} categories, respectively. The data statistics are shown in Table~\ref{tab:dataset}.


\begin{table}[!t]
\centering
\caption{ Datasets Statistics. Guardian News has the largest average document length and the most imbalanced class distribution.}\small
\begin{tabular}{lccll}
\toprule
\large
\textbf{Datasets} & \textbf{Avg. Length} & \textbf{Class ratio} &\textbf{\#Train}&\textbf{\#Test} \\[6pt] \hline
\textbf{Yelp} & 139& 2:1 & 140k&20k \\[5pt]
\textbf{IMDB} &218& 1:1 &25k& 25k\\ [5pt]
\textbf{Guardian} & 1,024& 4:2.5:2:1:0.5 & 37.03k&15.87k \\
\bottomrule
    \end{tabular}
    \label{tab:dataset}
\end{table}

\subsection{Baselines}

We compare our approach with the following baselines:
\begin{itemize}
\item{\emph{CNN}}: In our experiments, the kernel sizes are 3,4,5, and the number of kernels of each size is 100.

\item{\emph{LSTM}}: For each document, words are fed into LSTM sequentially and composed by mean pooling. \yh{A softmax layer is stacked to generate the class prediction}. We also report the results of LSTM with an attention mechanism (LSTM+Att).

\item{\emph{HAN}} \cite{DBLP:conf/naacl/YangYDHSH16}: \yh{The Hierarchical Attention Network which stacks two bidirectional Gated Recurrent Units and applies two levels of attention mechanisms at the sentence-level and at the document-level, respectively.} 

\item{\emph{BERT}} \cite{DBLP:conf/naacl/DevlinCLT19}: We feed each document into BERT as a long sequence with sentences separated by the \texttt{[SEP]} token, which is fine-tuned on our data. We truncate documents with length over 512 tokens and use the representation of the \texttt{[CLS]} token for classification. 

\item{\emph{Scholar}} \cite{DBLP:conf/acl/SmithCT18}: A neural topic model trained with variational autoencoder~\cite{DBLP:journals/corr/KingmaW13} with document-level class labels incorporated as supervised information. \YH{Scholar essentially learns a latent topic representation of an input document and then predicts the class label conditional on the latent topic representation}.

\item{\emph{VMASK}}~\cite{DBLP:conf/emnlp/ChenJ20}: \yh{The model applies} variational word masking \yh{strategy to mask out unimportant words to improve interpretability of neural}
text classifiers. \hqDec{During training, the binary mask is derived from the \textsl{Gumbel-softmax} operator on the non-linear transformation of an input sentence, and then element-multiplication is applied on the mask and the sentence to remove the unimportant words. In inference, they use softmax to get a softened version of the mask, instead. We report results from two variants of VMASK, by using the text input encoded either by BERT or LSTM.}
\footnote{Our results on IMDB are different from those reported in the original paper due to different train/test splits.}
\end{itemize}

\subsection{Data Pre-processing}

For the IMDB reviews, we use the processed IMDB dataset provided by Scholar\footnote{ \href{https://github.com/dallascard/scholar/blob/master/download_imdb.py}{IMDB dataset download script.}}. For both review datasets, we set the maximum sentence length to $60$ words and the maximum document length to $10$ sentences. We only keep the most frequent $15,000$ words in the training set, and mark the other words as $[\rm{unk}]$. Sentences with more than 30\% $[\rm{unk}]$ are removed from our training set. For the Guardian news data, we download the dataset from Kaggle\footnote{ \href{https://www.kaggle.com/sameedhayat/guardian-news-dataset/task}{Guardian News dataset.}} and follow its provided train/test split. We set the maximum sentence length to $60$ words and the maximum document length to $18$ sentences.

\section{Experimental Results}

\subsection{Text Classification Results}

The text classification 
results are shown in Table~\ref{tab:main_results}. Methods marked with $\dagger$ are re-implemented by us. 
As shown in Table~\ref{tab:main_results}, the vanilla classification models, such as CNN and LSTM, show inferior performance across three datasets. With the incoporation of the attention mechanism, LSTM-att slightly improves over LSTM. HAN was built on bidirectional GRUs but with two levels of attention mechanism at the word- and the sentence-level. It outperforms LSTM-att. BERT was built on the Transformer architecture. But it gives slightly worse results compared to HAN. 
The hierarchical modelling in HAN may explain its comparatively superior performance. The neural topic modelling approach, Scholar, performs better than CNN, but slightly worse than other baselines on IMDB and Yelp. 
VMASK 
learns to assign different weights to word-level features by minimising the classification loss. \YH{Its BERT variant generally outperforms the LSTM variant and gives the best results among the baselines. We have additionally performed statistical significance test, \hqMay{the Student's $t$-test}, to compare the performance of \textsc{Hint} with VMASK-BERT \hqMay{by training both models for 10 times}, and showed the results in Table~\ref{tab:main_results}.} 
In general, \textsc{\YH{Hint}} outperforms all baselines and the improvement is more prominent on the largest Guardian News dataset with longest average document length. 

\begin{table}[!t]
\centering
    \caption{ Classification accuracy on the three datasets. \YH{** significant at $p<0.05$, *** significant at $p<0.001$.}}\small
    \begin{tabular}{llll}
\toprule[1pt]
    Methods& IMDB & Yelp &Guardian  \\ \hline
    CNN$\dagger$
    & 83.36&94.16&92.82\\
    LSTM$\dagger$~
    &87.30&97.10&93.57\\
    LSTM-att$\dagger$&87.56&97.30&93.97\\
    HAN~
    &87.92&97.70&94.34\\
    BERT~
    &87.59&97.52&94.28\\
    Scholar~
    &86.10&96.87&93.97\\
    VMASK\hqDec{-BERT}~
    &{88.23}***&{98.10}**&\text{94.49}**\\
    \hqDec{VMASK-LSTM}~
    &87.40&98.04&93.79\\
    \textsc{\YH{Hint}}&\textbf{89.11}***&\textbf{98.42}**&\textbf{95.38}**\\
    \bottomrule
    \end{tabular}
    \label{tab:main_results}
\end{table}

\begin{figure}[!t]
    \centering
     \includegraphics[width=0.85\textwidth,trim={0.35cm 0.2cm 0.2cm 0.5cm},clip]{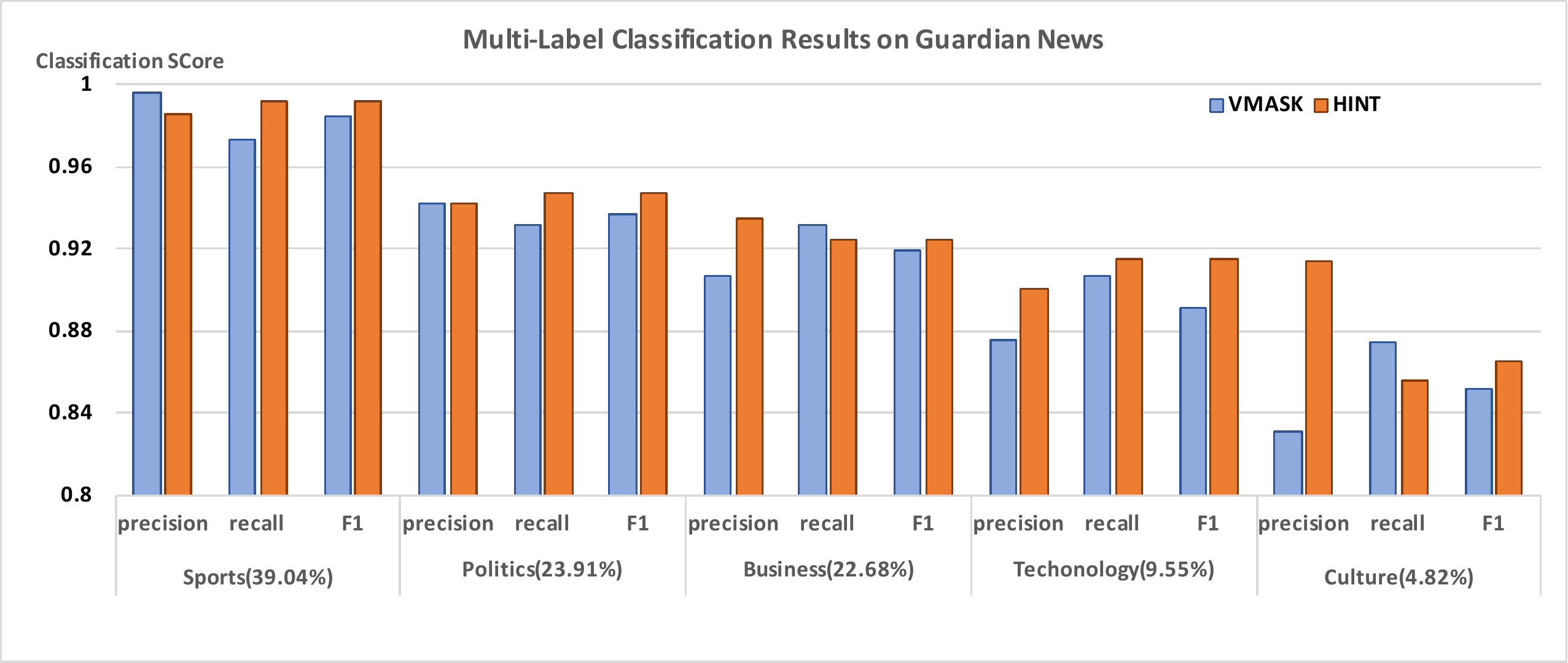}   
    \caption{ The precision, recall, and F1 of per-class classification results of VMASK and \textsc{\YH{Hint}} on the Guardian News dataset.}
    \label{fig:news_f1}
\end{figure}

To further examine the ability of \textsc{\YH{Hint}} in dealing with imbalanced data, we plot in Figure~\ref{fig:news_f1} the per-class precision, recall and F1 results on the Guardian News data. While \textsc{\YH{Hint}} generally outperforms VMASK in F1 across all classes, it achieves much better results on minority classes. For example, \textsc{\YH{Hint}} improves upon VMASK by nearly 8\% in precision on the smallest \emph{Culture} class.

\subsection{Topic Evaluation Results}


The \textit{sentence-level topic representation learning module} in the \textsc{\YH{Hint}} framework generates latent topic vectors which allows us to extract top associated words for each latent topic dimension by the weights connecting between the latent topic vector with the reconstruction layer. \hqMay{Existing work shows that a good latent variables should be able to
cluster the high-dimensional text representations into coherent
semantic groups~\cite{DBLP:journals/corr/KingmaW13}}. \hqMay{As described in Section~\ref{sec:generate_interpretation},} we can interpret the top associated words for each latent topic dimension as topic words.  
In this subsection, we show the topic extraction results \hqMay{by displaying the top 10 words in each latent dimension as word cloud.} 

\begin{figure}[!t]
    \centering
   \begin{subfigure}{.5\textwidth}
  \centering
   \includegraphics[width=0.9\textwidth,trim={0cm 3cm 2cm 1cm},clip]{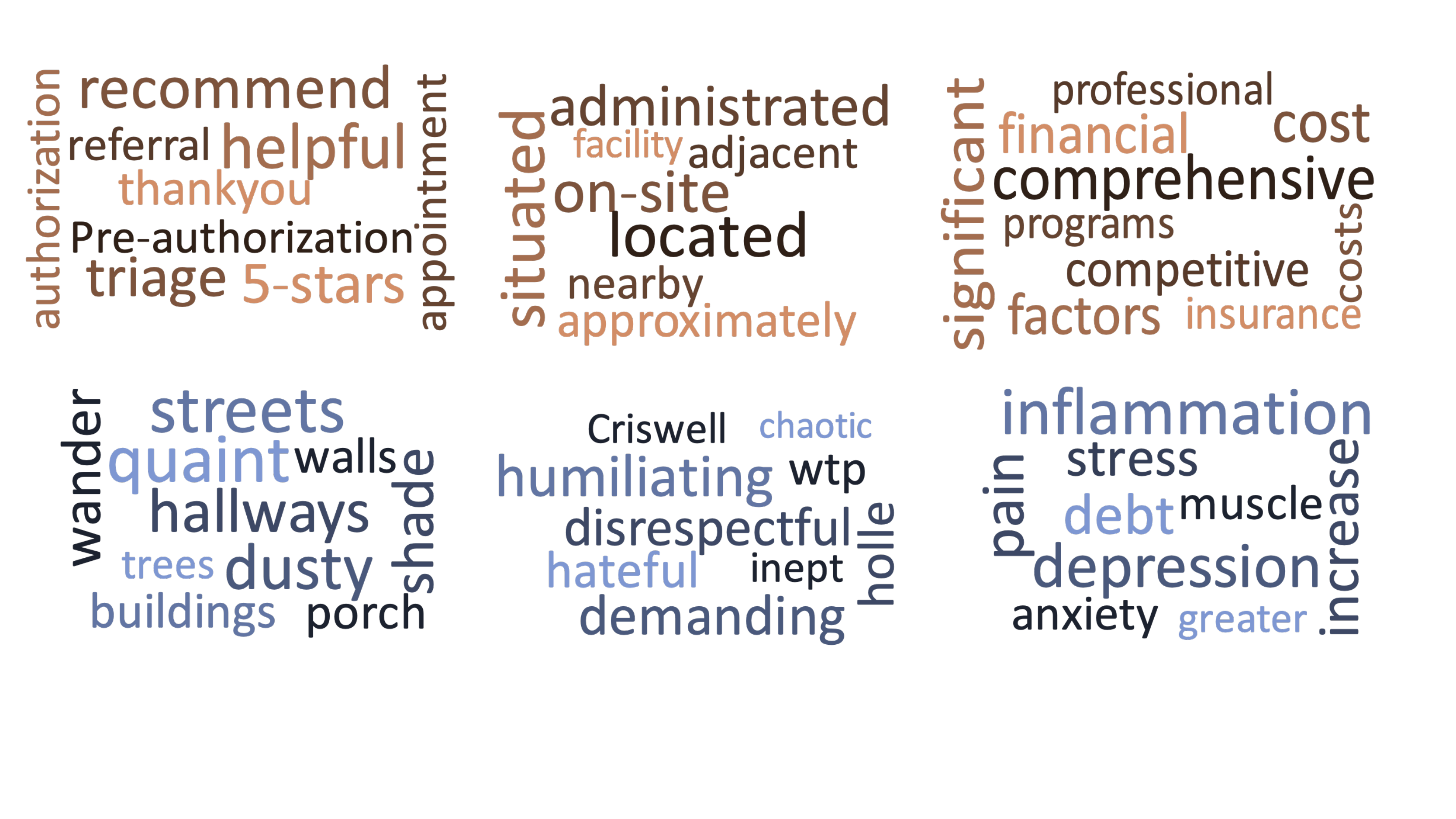}
  \caption{ Example topic words from Yelp.}
  \label{fig:review_wordcloud}
    \end{subfigure}%
    \hfill
    \begin{subfigure}{.5\textwidth}
    \centering
    \includegraphics[width=0.9\textwidth,trim={1cm 2.5cm 1.5cm 2.2cm},clip]{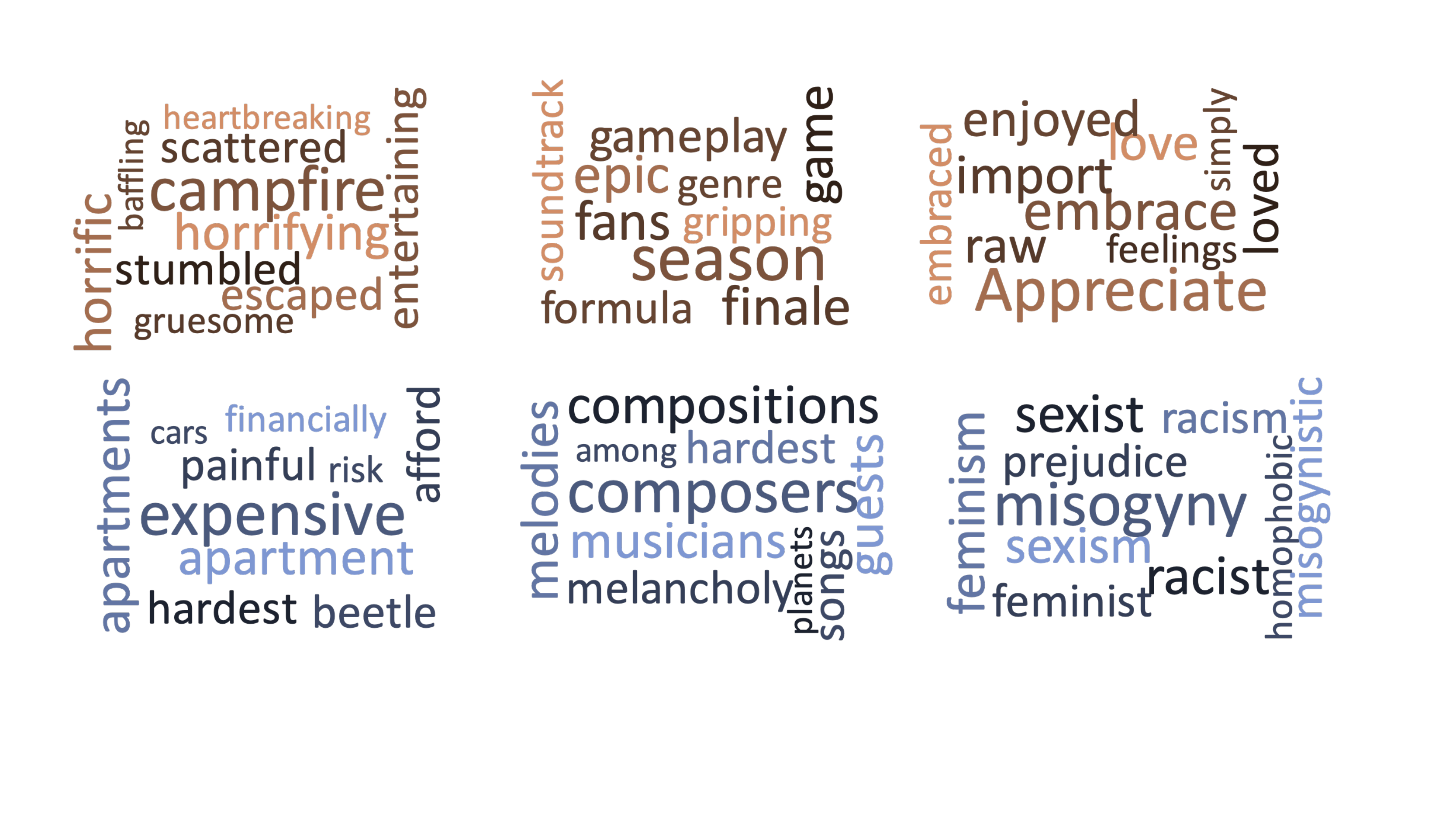}
    \caption{ Example topic words from IMDB.}
    \label{fig:imdb_wo}
    \end{subfigure}%
    \caption{ The topic word importance is indicated by the word font size. In each sub-figure, the upper topic word clouds are Positive, while the lower topic word clouds are Negative.}
\end{figure}

\begin{figure}[!t]
    \centering
    \includegraphics[width=0.7\textwidth,trim={1cm 1.5cm 1.5cm 2.2cm},clip]{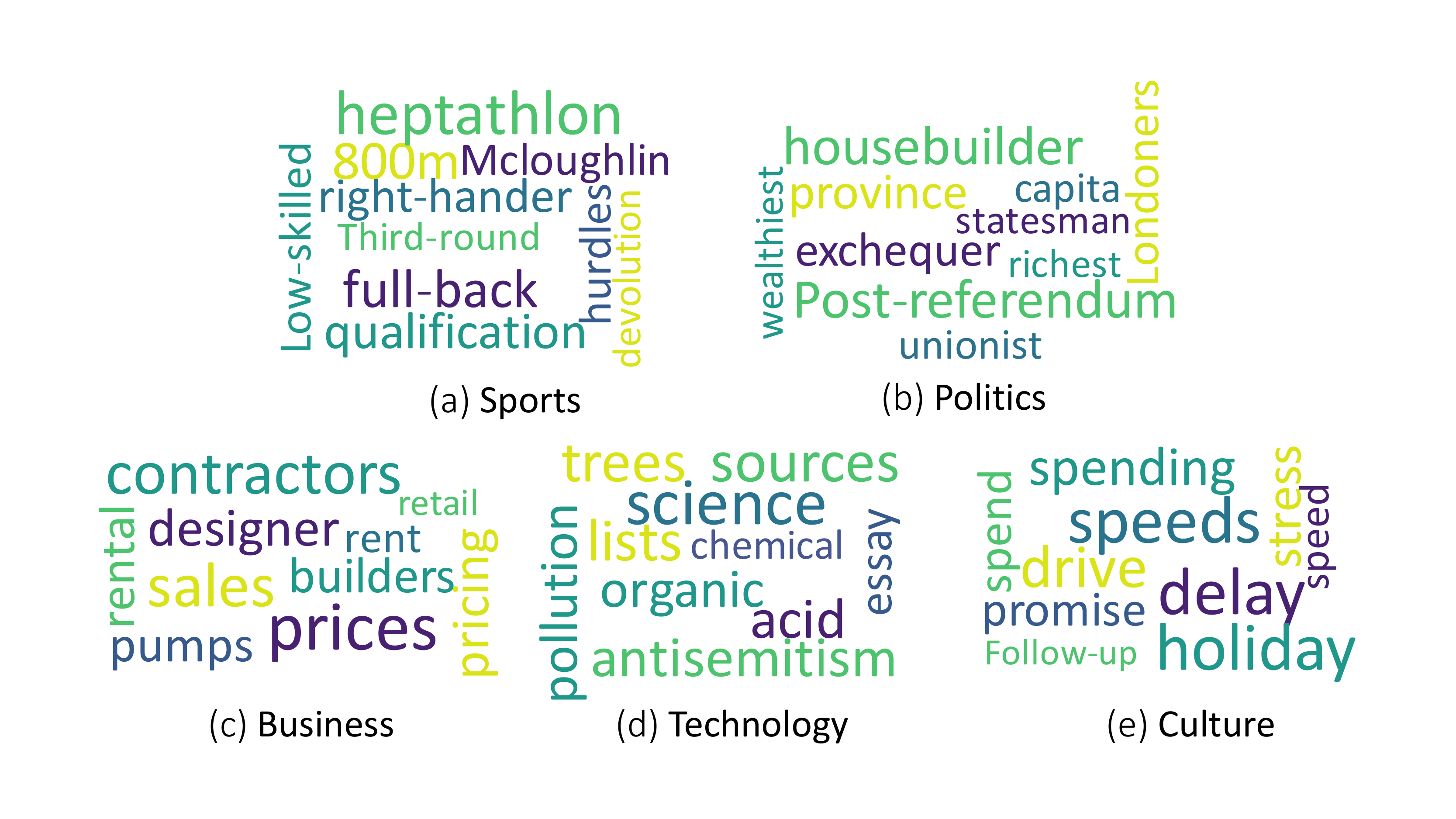}
    \caption{ Example topic words for news categories in the Guardian News Dataset.}
    \label{fig:news_wo}
    \vspace{-0.3cm}
\end{figure}

Figure~\ref{fig:review_wordcloud} and \ref{fig:imdb_wo} show the word clouds of the generated example topic words on the Yelp and the IMDB, respectively. 
\yh{It can be easily inferred from Figure~\ref{fig:review_wordcloud} that users express general positive comments, praise convenient facility locations and competitive pricing; while they complain about dusty environment, service quality and express negative feeling relating to their diseases. } 
In Figure~\ref{fig:imdb_wo}, we can observe that reviewers' attitudes towards different genres of movies. They like \emph{thriller} and \emph{animated movie}. On the contrary, they show negative feelings towards luxury \emph{lifestyle} or movies relating to \emph{misogyny}. These results show that \textsc{\YH{Hint}} can indeed extract topics discussed under different polarity categories despite using no topic-level polarity annotations for topic learning. Figure~\ref{fig:news_wo} shows example topic words each of which corresponds to the five news categories from the Guardian news data. 

\hqMay{In addition to visualising the extracted topics, we also}
 evaluate the quality of the extracted topics using four different topic coherence measures, including the normalised Pointwise Mutual Information (NPMI), a lexicon-based method (UCI), and context-vector-based coherence measures (CV). We compare the results with LDA \cite{blei2003latent} and Scholar~\cite{DBLP:conf/acl/SmithCT18} in Table~\ref{tab:topic_coherence}. It can be observed that overall, \textsc{\YH{Hint}} gives the best results on Yelp. It performs worse than Scholar on the Guardian News in UCI, but achieves better results in CV and NPMI. On the IMDB dataset, however, \textsc{\YH{Hint}} only outperforms the other two models in CV and was beaten by LDA in both NPMI and UCI. One possible reason is that \textsc{\YH{Hint}} estimates the word probability by context embedding. Hence, it beats the baselines on context-vector-based coherence, but only achieves comparable performance on the lexicon-based metric.
 
 \begin{table}[!t]
 \centering
     \caption{ Topic Coherence results.}\small
     \begin{tabular}{lccccccccc}
 \toprule[1pt]
 \multirow{2}{*}{\textbf{Method}} & \multicolumn{3}{c}{\textbf{IMDB}} & \multicolumn{3}{c}{\textbf{Yelp}} & \multicolumn{3}{c}{\textbf{Guardian News}} \\
 \cmidrule(lr){2-4} \cmidrule(lr){5-7} \cmidrule(lr){8-10}
 & CV&NPMI&UCI &CV&NPMI&UCI&CV&NPMI&UCI\\
     \midrule
LDA &0.341&\bf-0.032&\bf-1.936 & 0.377&\bf-0.039&-1.495 &0.362&-0.140&-2.858\\
 Scholar &0.351&-0.057&-2.010& 0.424&-0.061&-2.188 &0.373&-0.286&\bf-1.207\\
 \textsc{\YH{Hint}} &\bf0.401&-0.068&-1.992 & \bf0.445&\bf-0.039&\bf-1.385 &\bf0.423&\bf-0.108&-3.085\\
\bottomrule
     \end{tabular}
     \label{tab:topic_coherence}
 \end{table}

\subsection{Interpretability Evaluation}
\yh{A good interpretation method} 
should give explanations that are (i) easily understood by humans (ii) indicative of \textit{true} importance of input features. 
\yh{We conduct both quantitative \YH{and human} evaluations on the interpretation results generated by \textsc{\YH{Hint}}.} 

\subsubsection{Word Removal Experiments}
\label{sec:boxplot}

\yh{A good interpretation model should be able to identify} truly important features when making predictions~\cite{alvarez2018towards}. A common evaluation \yh{strategy} 
is to remove features \yh{identified by the interpretation model}, and measure the drop in the classification accuracy~\cite{DBLP:conf/emnlp/ChenJ20}. 

\begin{figure}[!t]
    \centering
    \includegraphics[width=0.8\textwidth,trim={1cm 2.8cm 2cm 1cm},clip]{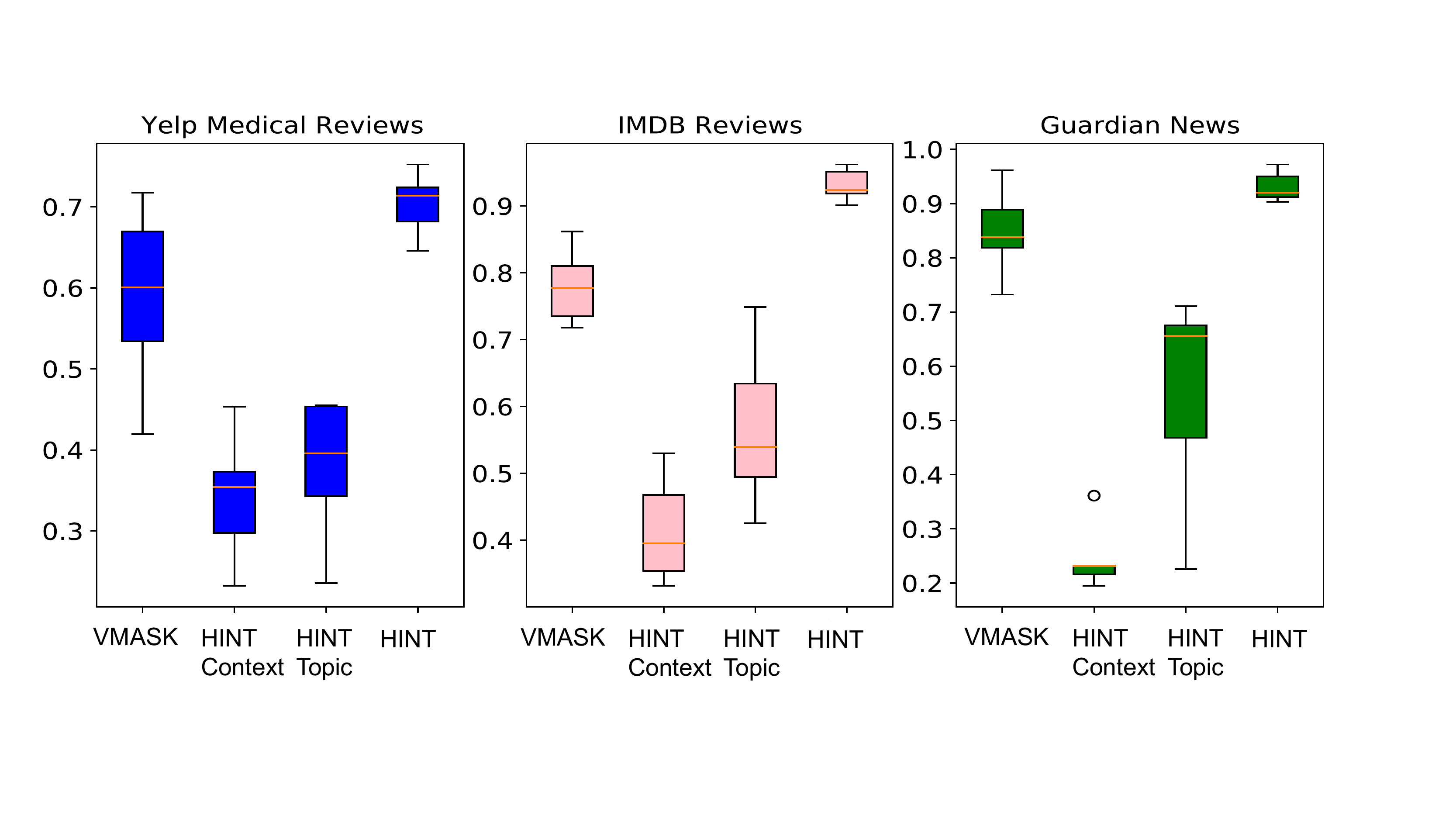}
    \caption{ Aggregated correlation score between the classification accuracy drop and the number of removed words. \textsc{\YH{Hint}} shows the highest correlation score, i.e., removing the top topic words identified by \textsc{\YH{Hint}} leads to more significant performance drop compared to \yh{word masking methods}.}
    \label{fig:faithfulness}
\end{figure}

Figure~\ref{fig:faithfulness} shows the correlation score between the accuracy drop and the number of removed words evaluated on the three datasets. In addition to VMASK, we also take two variants \yh{of \textsc{\YH{Hint}}} as the contrasts, i.e., \textsc{\YH{Hint}}-Context and \textsc{\YH{Hint}}-Topic. The former masks the words assigned with large $\alpha_{ij}$ weights by the context learning module, the latter masks the words with large $\beta_{ij}$ attention weights in \yh{the topic learning module}. 
\textsc{\YH{Hint}} masks the top-$\textsc{K}$ \yh{unique} topic words according to their weights in each topic. \yh{It can be observed that simply masking words with higher weights identified by the context learning module or the topic learning module does not give good correlation scores. The results are worse than VMASK which automatically determine which words to mask based on the information bottleneck theory. Nevertheless, when masking words based on those identified by \textsc{\YH{Hint}}, we observe better correlation scores with smaller spreads compared to VMASK, showing the effectiveness of \textsc{\YH{Hint}} in identifying task-important words.} 

In Table~\ref{tab:top}, we list part of top $k$ words removed by \textsc{\YH{Hint}} and VMASK on the IMDB dataset with the corresponding performance drops. 
It can be observed that when $k$ is 20, both methods tend to identify opinion words as key features for removal which are task-relevant, resulting in a similar classification accuracy drop. With the increasing number of $k$, VMASK still primarily focuses on opinion words, which leads to a further modest accuracy drop. On the contrary, when $k$ increases, \textsc{\YH{Hint}} started to extract topic-related words such as `\emph{comedies}' `\emph{screenwriter}'. These words may seem to be task-irrelevant. However, the removal of them causes more noticeable accuracy drop. We speculate that such words are highly relevant with the latent topics discussed in text, which in turn are associated with implicit polarities important for the decision of document-level sentiment classification. 

\begin{table}[!t]  
\centering
    \caption{ Accuracy drop by VMASK and \textsc{\YH{Hint}} with different number of removed words. }\small
    \begin{tabular}{cccm{7cm}} \toprule
         \textbf{k} & \textbf{ACC$\downarrow$} & \textbf{Method} & \textbf{Removed words}  \\\midrule
          \multirow{2}{*}{20}
          &1.1\%& VMASK &brilliantly best intelligently tough cynicism\\ 
          &1.2\%&  \textsc{\YH{Hint}} &  unwatchable highest dramas deeply flawless \\ \midrule 
          \multirow{2}{*}{40}
          &2.8\%& VMASK & interesting timeless lacks failed remaining  \\ 
          &3.8\%&  \textsc{\YH{Hint}}  & perfection comedies screenwriter reporter disagree\\ \midrule 
          \multirow{2}{*}{60}
          &2.3\%& VMASK &recommend like pretty suggestion poorly
          \\ 
          &4.6\%&  \textsc{\YH{Hint}} & scripts mysteries complaint funeral werewolf \\ 
          \bottomrule
    \end{tabular}
    \label{tab:top}
\end{table}

\subsubsection{Human Evaluation for Interpretablity}
\hqDec{We conduct human evaluation to validate the interpretablity of our proposed method on the following criteria inspired by existing methods on human evaluation~\cite{DBLP:conf/nips/ZhouHZLSXT20}:}
\begin{itemize}
    \item \emph{Correctness}. \hqDec{It measures to what extend users can make correct prediction given the model interpretations. That is, users are asked to predict the document label based on the model-generated interpretation. If the interpretation is correct, then users should be able to predict the document label easily.}
    \item \emph{Faithfulness.} \hqDec{It measures
to what extent the generated explanation is faithful to the model prediction.}
\item \emph{Informativeness}. \hqDec{It measures to what extend the interpretation reveals the key information conveyed in text such as the main topic discussed in text, its associated polarity, and the secondary topic (if there is any) mentioned in text. }
\end{itemize}

\begin{figure}[!t]
\centering
\includegraphics[width=0.85\textwidth,trim={90 250 90 95},clip]{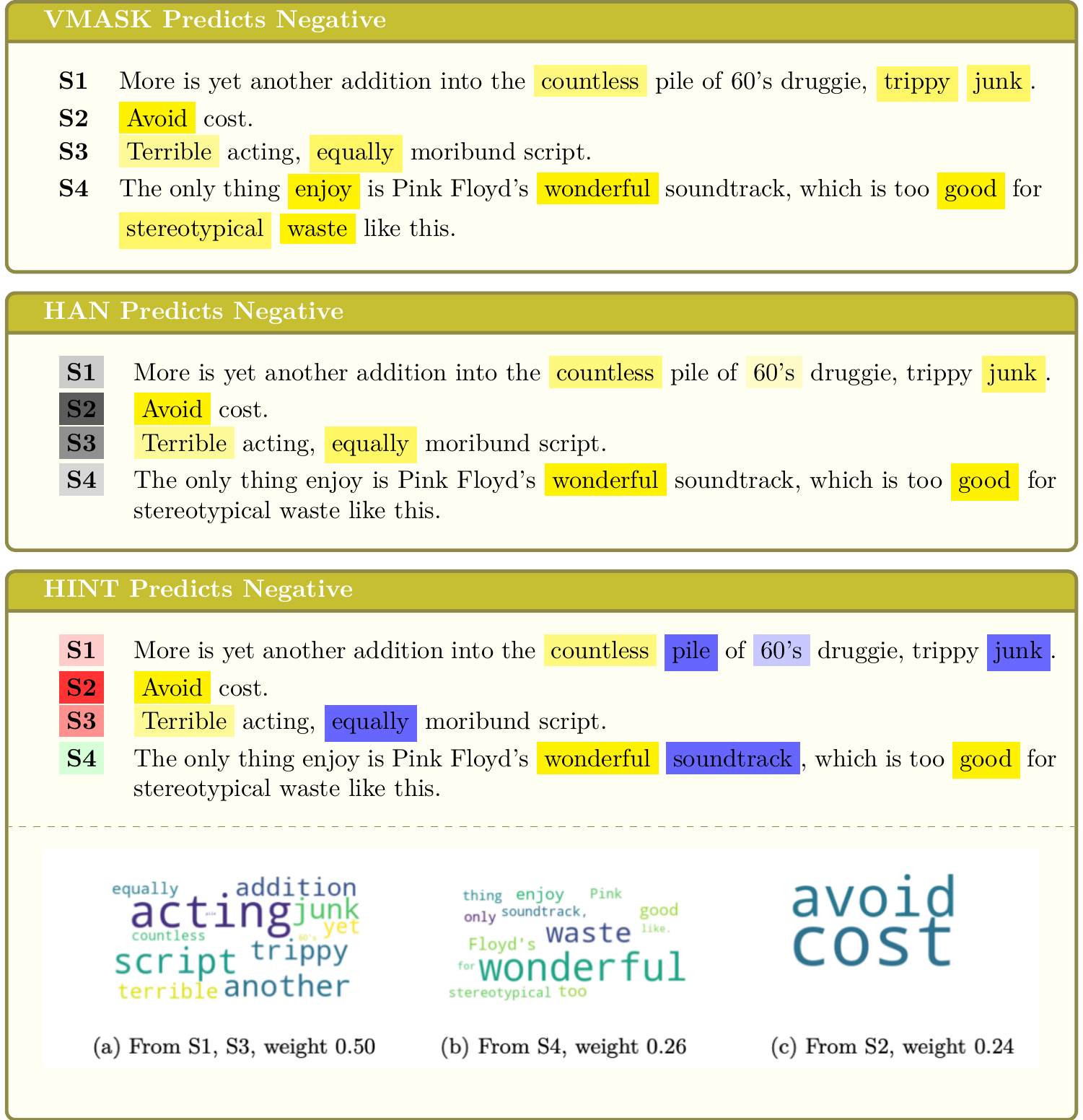}
\caption{ \hqMay{Interpretations generated by VMASK, HAN and \textsc{Hint} on the same \textsf{IMDB} review with mixed sentiments. VMASK only highlights important words for classification, while HAN additionally displays sentence importance. 
The interpretations generated by \textsc{Hint} contain richer information. It highlights both label-independent words (in blue) and label-dependent words (in yellow) as the word-level interpretations; it also displays the sentiment strength (red for negative, green for positive) for each sentence as the sentence-level interpretations. Moreover, \textsc{Hint} groups sentences into three topics (shown as three word clouds) with the document-level topic weights. Along with the sentence-level sentiment, we can easily tell that the document contains mixed sentiments, and the most predominant topic (associated with S1, S3) is negative, thus inferring the document as negative.}}
\label{fig:human_evaluation_case}
\vspace{-0.5cm}
\end{figure}

\hqDec{We randomly select 100 samples  with the interpretations generate by HAN~\cite{DBLP:conf/naacl/YangYDHSH16}, VMASK~\cite{DBLP:conf/emnlp/ChenJ20} and our model for evaluation. We invite three evaluators, all are proficient in English and with at least MSc degrees in Computer Science, to score the interpretations generated on the sampled data in a likert scale of 1 to 5. Details of the evaluation protocol are presented in Appendix A.}

\hqMay{We show interpretations generated from each of the three models in Figure~\ref{fig:human_evaluation_case} for a movie review with mixed sentiments. The review expresses a negative polarity towards the topic of \emph{acting} and \emph{script}, while a positive polarity for the \emph{soundtrack}, resulting in an overall negative sentiment. For \emph{Correctness}, the evaluators are required to predict the document label only based on the generated interpretations 
without reading the document content in detail. We can observe that VMASK highlights both positive and negative words, e.g., `\emph{junk}' and `\emph{enjoy}', making it relatively difficult to infer the document-level sentiment label. HAN additionally provides the sentence-level importance from which we know that sentence S2 is more important than the others and it contains the negative word `\emph{avoid}'. Compared to the baselines, \textsc{Hint} reveals much richer information. One can easily tell that the document contains mixed sentiments as the first three sentences carry a negative sentiment while the last one bears a positive polarity. In addition, the document discusses three topics (shown as three word clouds) with S1 and S3 associated with the most prominent topic about \emph{acting} and \emph{script}, carrying a negative sentiment. Thus, it seems that the \textsc{hint}-generated interpretations align with the model-predicted label (i.e., \emph{Faithfulness}) and it also provide a higher level of \emph{Informativeness}. 
The human evaluation results are shown in Table \ref{tab:humanEvaluation}. It can be observed that \textsc{Hint} gives the best results among all criteria.}

\renewcommand{\arraystretch}{1.2}
\begin{table}[!t]
    \centering\small
    \begin{tabular}{lccc}
\toprule[1pt]
Model& \textbf{Correctness} &\textbf{Faithfulness} & \textbf{Informativeness} \\ 
\midrule
HAN& 3.89 & 3.92& 3.79 \\ 
VMASK & 4.13 & 4.06 & 3.93 \\ 
\textsc{Hint} & \textbf{4.37} & \textbf{4.28} & \textbf{4.11} \\ 
\bottomrule[1pt]
    \end{tabular}
    \caption{     Human evaluation results in a likert scale of 1 to 5 (1: \emph{Strongly Disagree}; 5: \emph{Strongly Agree}). The inner-rater agreement measured by Kappa score is 0.37.}
    \label{tab:humanEvaluation}
\end{table}

\subsubsection{Completeness and Sufficiency on ERASER}
\hqDec{We also use the Evaluating Rationales And Simple English Reasoning (\textsf{ERASER}) benchmark~\cite{DBLP:conf/acl/DeYoungJRLXSW20} to evaluate the model interpretability. \textsf{ERASER} contains seven datasets which are repurposed from existing NLP corpora originally used for sentiment analysis, natural language inference, Question-Answering, etc. Each dataset is augmented with human annotated rationales (supporting evidence) that support output predictions. We select \textsf{Movie Reviews}\footnote{ \url{http://www.eraserbenchmark.com/zipped/movies.tar.gz}} as our evaluation dataset. In ERASER, \textsf{Movie Reviews} only contains a total of 1600 documents, another 200 test samples have been annotated with human rationales which are text spans indicative of the document polarity labels. We train all the models on our IMDB dataset and evaluate on the annotated \textsf{Movie Reviews}.}

\begin{figure}[!t]
    \centering
    \includegraphics[width=0.98\textwidth]{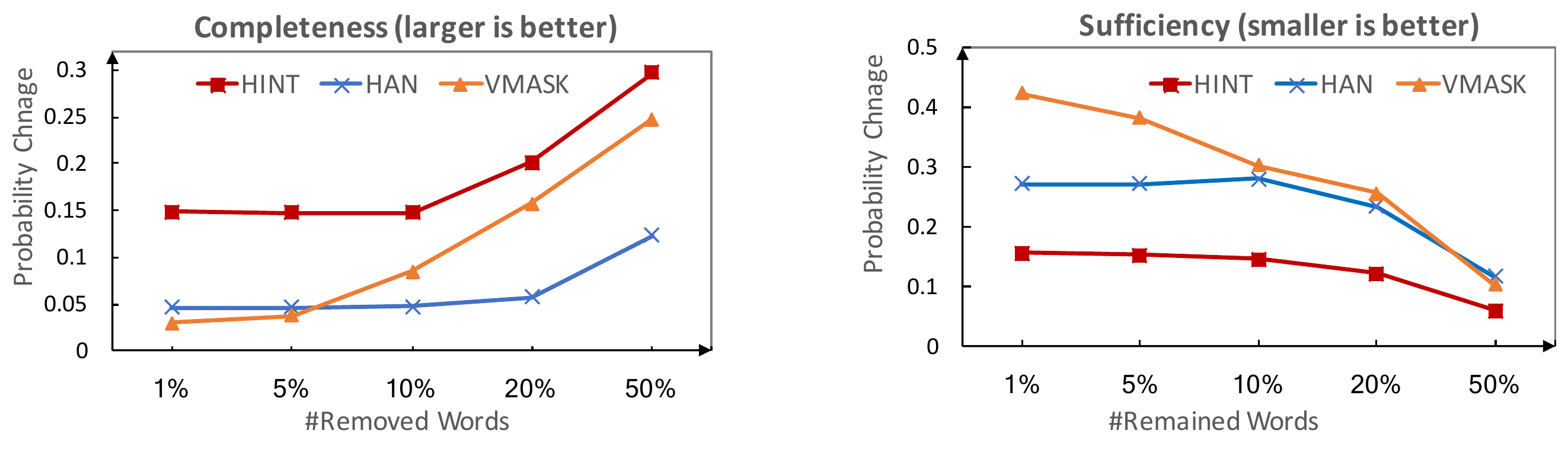}
    \caption{ \YH{The \textsl{Completeness} and \textsl{Sufficiency} values by removing or keeping the top $k\%$ most important word tokens identified by various models, $k\in \{1, 5, 10. 20, 50\}$. } }
    \label{fig:Comp_Suff}
\vspace{0cm}
\end{figure}

Following what has been proposed in \textsf{ERASER}, we first evaluate model interpretation using the two metrics, \underline{\emph{Completeness} and \emph{Sufficiency}}, which measure the model prediction changes after removing the important words identified by the model and merely based on the important words\footnote{ \hqMay{For \textsc{hint}, we select the important words according to their weights from both the context representation learning and the topic representation modules.}}, respectively. That is:
\begin{eqnarray}
    \text{completeness} &=& m(x_{i})_{j}-m(x_{i}/e_{i})_{j},\\
    \text{sufficiency} &=& m(x_{i})_{j} -m(e_{i})_{j},
\end{eqnarray}
where $x_{i}$ is the original text, $e_{i}$ is the identified important words, $m(\cdot)$ is the model probability on the predicted label $j$. To study the effects of word tokens in different importance level, we follow the setup in ERASER and group word tokens into 5 bins, each corresponding to the top 1\%, 5\%, 10\%, 20\%, 50\% of most important tokens identified by a model. The results are shown in Figure~\ref{fig:Comp_Suff}. We can observe that on \emph{Completeness}, HAN and VMASK perform similarly with up to 5\% most important words removed. But with more words removed, VMASK gives superior performance compared to HAN. \textsc{Hint} outperforms both HAN and VMASK consistently, but with the performance gap reduced when removing more words from documents. On \emph{Sufficiency}, \textsc{Hint} and HAN give superior results compared to VMASK when only keeping a small number of important words. However, when over 20\% most important words are kept, the performance difference between HAN and VMASK diminishes. Overall, \textsc{Hint} gives the best results among all the models.



\subsubsection{Agreement with human rationales}
\label{sec:rationale_result}

We use the multiple-aspect sentiment analysis dataset, \textsf{BeerAdvocate}~\cite{mcauley2012learning}, consisting of beer reviews, each of which is annotated with five aspects and the aspect-level rating scores in the range of 0 to 5. 
It has been widely used in evaluating rationale extraction models~\cite{DBLP:conf/acl/BastingsAT19,DBLP:conf/emnlp/LeiBJ16,DBLP:conf/emnlp/LiE19,DBLP:conf/nips/YuZCJ21} by calculating the agreement between the annotated sentence-level rationales and model identified text spans. The common pipeline in many rationale extraction work is to predict binary masks for rationale selection, i.e., masking the unimportant text spans and then predicting the sentiment scores only based on the selected rationales. In the HardKuma approach proposed for rationale extraction, constraints are further imposed to guarantee the continuity and sparsity of the selected text spans~\cite{DBLP:conf/acl/BastingsAT19}. More recently, \citet{DBLP:conf/nips/YuZCJ21} argued that such a two-component pipeline approach tends to generate suboptimal results since even the first step of rationale selection selects a sub-optimal rationale, the sentiment predictor can still produce a lower prediction loss. To overcome this problem, they proposed the Attention-to-Rationale (A2R) approach by adding an additional predictor which predicts the sentiment scores based on soft attentions as opposed to the selected rationales. During training, the gap between the two predictors, one based on the selected rationales while the other based on soft attentions, is minimised. 
We show the results of both HardKuma and A2R reported in \cite{DBLP:conf/nips/YuZCJ21} in the upper part of Table~\ref{tab:rationale_extraction}\footnote{ Note that the HardKuma results reported in \cite{DBLP:conf/nips/YuZCJ21} are inferior than those reported in the original paper \cite{DBLP:conf/acl/BastingsAT19}. This is because the strong continuity constraint in HardKuma was not used in \cite{DBLP:conf/nips/YuZCJ21} in order to achieve a fair comparison with A2R.}.

In our experiments, we train \textsc{Hint} and the baselines, VMASK and HAN, on the \textsf{BeerAdvocate} training set, and stop training when the models reach the smallest Mean Square Error (MSE) on the validation set. Afterwards, rationale selection is performed based on the word-level attentions for VMASK (We use the top 15\% words as rationales), and based on the sentence-level importance scores for HAN and \textsc{Hint}. For the latter two models, we only extract the top sentence as the rationale for each document in the test set. 

Following the setting in~A2R \cite{DBLP:conf/nips/YuZCJ21}, the overlapping between the selected important words or sentences and the gold-standard rationales are calculated as precision and recall values and are shown in Table~\ref{tab:rationale_extraction}. It can be observed that approaches specifically designed for rationale extraction, HardKuma and A2R, give better results compared to other approaches which are not optimised for rationale extraction. VMASK performs the worst as it can only select token-level rationales. \textsc{Hint} outperforms HAN on both the \emph{Look} and the \emph{Smell} aspects by a large margin, and the two models give similar results on the \emph{Palate} aspect.


\begin{table}[!t]
    \centering
    \resizebox{0.85\textwidth}{!}{
    \begin{tabular}{lcccccc}
\toprule[1pt]
&\multicolumn{2}{c}{Look}&\multicolumn{2}{c}{Smell}&\multicolumn{2}{c}{Palate}\\
\cmidrule(lr){2-3} \cmidrule(lr){4-5} \cmidrule(lr){6-7}
& Precision & Recall & Precision & Recall  & Precision & Recall\\
\hline
HardKuma
 & 81.0 & 69.9 & 74.0 &\textbf{72.4} &  45.4& \textbf{73.0}   \\
A2R
 & \textbf{84.7}& \textbf{71.2}&  \textbf{79.3} &71.3& 64.2& 60.9   \\
\hline
VMASK
& 33.8 & 28.5 & 16.0 & 13.5 & 27.0 & 36.8  \\
HAN
& 76.1 & 58.2& 56.0 & 48.1 & \textbf{71.6} & 66.0   \\
\textsc{Hint} 
&84.4 & 67.0 & 59.4 & 54.8 & 70.4 & 65.1\\
\bottomrule[1pt]
    \end{tabular}
    }
    \caption{ Precision, recall of rationale extraction on the three aspects in the \textsf{BeerAdvocate} dataset. The results of HardKuma and A2R are taken from \cite{DBLP:conf/nips/YuZCJ21}.} 
    \label{tab:rationale_extraction}
\end{table}

\subsection{Ablation Study}

\hqDec{To study the effects of different modules in our model, we perform an ablation study and show the results in Table~\ref{tab:ablationStudy}. In addition to the accuracy on the three datasets, we also report the interpretability metrics, i.e., completeness and sufficiency for different variants~\footnote{ We randomly select 200 test samples to evaluate the interpretability.}. 
For \textsl{Topic Representation Learning} (\textsection\ref{sec:topic_learning}), we remove the Bayesian inference part which is used to learn word-level weight $\bm{\beta}_{ij}$. That is, rather than using $\bm{\beta}_{ij}$ to aggregate the word representations $x_{ij}$ in order to derive the sentence embedding $\bm{r}_{i}$ as shown in Figure~\ref{fig:sentTopicRepLearning}, we now derive the sentence embedding $\bm{r}_{i}$ using the word-level TFIDF weights to aggregate the word representations $x_{ij}$. 
We also explore the effects without the regularisation terms defined in Eq. \ref{eq:topicOrthogonal} and \ref{eq:disc}, respectively. Finally, we study the impact with or without the Graph Attention Networks (GATs) and the number of GATs layers in \emph{Document Representation Learning} (\textsection\ref{sec:doc_graph}). From the results in Table~\ref{tab:ablationStudy}, \textsc{Hint} achieves best performance on accuracy and overall better performance on interpretability metrics. The variant of using uniform weight as an initialisation for topic learning shows good interpretability on IMDB. This shows that with our proposed stochastic learning process for topic-related weights, it does not matter whether the word token weights are initialised by TFIDF or a uniform distribution. Although using multiple GAT layers fails to bring improvement to classification accuracy, 4-layer GAT has overall better interpretability performance than other GAT configurations.}



\begin{table}[!t]
    \centering
    \resizebox{0.98\textwidth}{!}{
    \begin{tabular}{lccccccccc}
\toprule[1pt]
\multirow{2}{*}{Methods}  & \multicolumn{3}{c}{IMDB} & \multicolumn{3}{c}{Yelp} & \multicolumn{3}{c}{Guardian}  \\
\cmidrule(lr){2-4} \cmidrule(lr){5-7} \cmidrule(lr){8-10}
&Acc($\uparrow$)&Com($\uparrow$)&Suff($\downarrow$)&Acc($\uparrow$)&Com($\uparrow$)&Suff($\downarrow$)&Acc($\uparrow$)&Com($\uparrow$)&Suff($\downarrow$)\\
 \midrule
\textsc{Hint} &\textbf{89.11}&\textbf{0.21}&0.11& \textbf{98.52}&\textbf{0.22}&0.09& \textbf{95.37}&\underline{0.16}&\underline{0.05}\\
 \midrule
Remove Bayesian inference for $\beta$ learning & 89.02&0.17&0.14&98.45&0.16& 0.11&95.21&\underline{0.16}&0.06\\
Replace TFIDF with uniform weight & 88.62&\underline{0.20}&\underline{0.10}&98.31 &0.12&0.09&95.08&\textbf{0.18}&0.06
\\
\midrule
w/o the Regularisation Term 1 (Eq. \ref{eq:topicOrthogonal})&89.03 &0.18&0.11&98.49&0.11&\textbf{0.07}&95.20&0.15&0.07\\
w/o the Regularisation Term 2 (Eq. \ref{eq:disc}) &89.06 &0.19&0.13&98.50&0.13&\underline{0.08}&95.26&0.13&0.08\\
\midrule
Remove GAT  &89.00&0.17&\underline{0.10}&98.41&\textbf{0.22}&0.10&94.87&0.08&\textbf{0.04}\\
2-layer GAT &88.93 &0.17&0.11&98.52&\underline{0.18}&0.10&94.99&0.14&0.06\\
4-layer GAT & 88.85&0.18&\textbf{0.07}&98.50 &\underline{0.18}&0.09& 93.58&0.13&\textbf{0.04}\\
\bottomrule[1pt]
    \end{tabular}
    }
    \caption{ Ablation study results showing the effects of different input for topic learning reconstruction, regularization term and GAT layers. Best accuracy and interpretability are marked \textbf{in bold}, the second best interpretablity is marked with \underline{underline}. \textsc{Hint} achieves the overall best results on most metrics.} 
    \label{tab:ablationStudy}
\end{table}





\subsection{Case Study}
\begin{figure}[!t]
\centering
\includegraphics[width=0.85\textwidth,trim={280 100 230 100},clip]{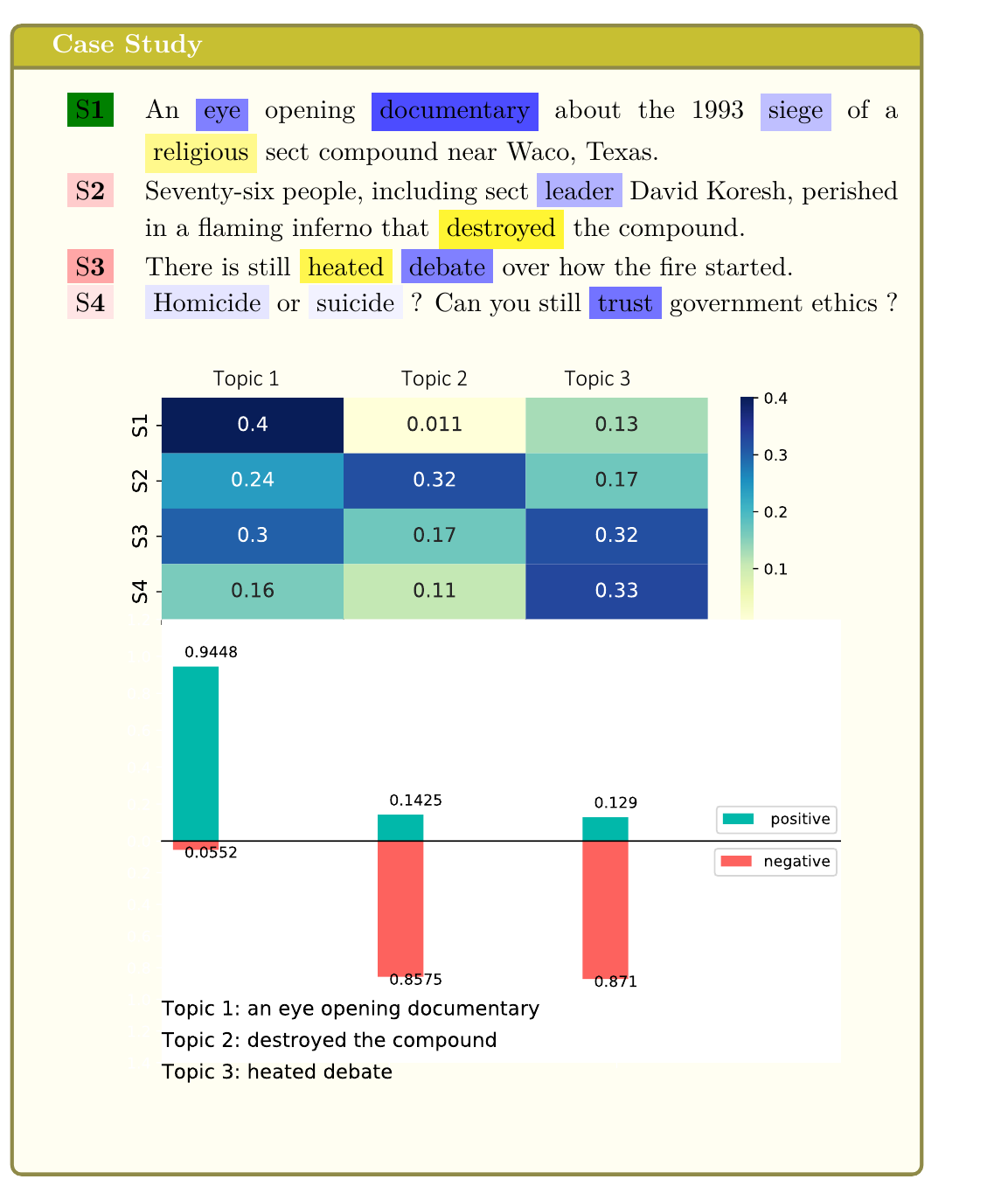}
\caption{ \hqMay{The upper part shows the document content with word-level important words and sentence-level sentiment labels. The middle and the lower parts show the topic-related interpretations generated by \textsc{Hint}. The heatmap shows the sentence-topic associations with a darker value indicating a stronger association, while the bar chart shows the sentiment strengths of the topics with the green and the red colours for the positive and the negative sentiment respectively. We also display the topic labels (shown under the bar chart) by automatically extracting the most relevant text span in the document to represent each topic. }}
\label{fig:case_study}
\vspace{-10pt}
\end{figure}

\hqDec{To show the capability of \textsc{Hint} in dealing with documents with mixed sentiments, we select one document from the IMDB dataset} \hqMay{to illustrate the interpretations generated 
in Figure~\ref{fig:case_study}. The figure consists of three parts. The top part shows the word-level interpretations in the form of label-dependent words (in yellow) and label-independent words (in blue), as well as the sentence-level sentiment labels (sentence IDs highlighted with red or green colours). The middle part shows a heat map illustrating the association strengths between sentences and topics with a darker value indicating a stronger association. The lower part presents a bar chart showing the sentiment strength of each topic with the green and the red colour for the positive and the negative sentiment respectively. To make it easier to understand what each topic is about, we automatically extract the most relevant text span in the document to represent each topic (shown under the bar chart) by the approach described in Section 4. } 

\hqMay{Our model derives the document label by aggregating the sentence-level context representations weighted by their 
topic similarities (see in \textsection\ref{sec:doc_graph}). From Figure \ref{fig:case_study}, we can observe that Topic 1 appears to be the most prominent topic in the document from the sentence-topic heat map. Sentence S2 is related to Topic 2, while both sentences S3 and S4 are grouped under Topic 3. Among the three topics, Topic 1 is positive, while Topic 2 and Topic 3 are negative. After aggregating sentences weighted by their topic similarities, the model infers an overall positive sentiment since the most prominent Topic 1 is positive.} \hqMay{This example shows that \textsc{Hint} is able to capture both the topic and sentiment changes in text.}

 
\section{Conclusion and Future Work}
\label{conclusion}
In this paper, we have proposed a \YH{Hierarchical} Interpretable Neural Text classifier, called \textsc{\YH{Hint}}, \yh{which automatically generates \YH{hierarchical} interpretations of text classification results}. It learns the sentence-level context and topic representations in an orthogonal manner in which the former captures the label-dependent semantic information while the latter encodes the label-independent topic information shared across documents. The learned sentence representations are subsequently aggregated by a Graph Attention Network to derive the document-level representation for text classification.  
We have evaluated \textsc{\YH{Hint}} on both review data and news data and shown that \yh{it achieves text classification performance on par with the existing neural text classifiers and generates more faithful} 
interpretations as verified by both quantitative and qualitative evaluations. 



While we only focus on interpreting neural text classifiers here, the proposed framework can be extended to deal with other tasks such as content-based recommendation. In such a setup, we will need to learn both user- and item-based latent interest factors by analysing reviews written by users and those associated with particular products. Since the proposed \textsc{\YH{Hint}} is able to extract topics and their associated polarity strengths from reviews, it is possible to derive user- and item-based latent interest factors based on the outputs produced by \textsc{\YH{Hint}}. \YH{Moreover, many NLP tasks such as natural language inference, rumour veracity classification, extractive question-answering and information extraction can be framed as classification problems. The proposed framework has a great potential to be extended to a wide range of NLP tasks.}



\section*{Acknowledgement}
This work was funded by the UK Engineering and Physical Sciences Research Council (grant no. EP/T017112/1, EP/V048597/1). HY receives the
PhD scholarship funded jointly by the University
of Warwick and the Chinese Scholarship Council. YH is supported by a Turing AI Fellowship funded by the UK Research and Innovation (grant no. EP/V020579/1).


\starttwocolumn

\bibliography{compling_style}

\begin{thebibliography}{58}
\expandafter\ifx\csname natexlab\endcsname\relax\def\natexlab#1{#1}\fi

\bibitem[{Abdou et~al.(2020)Abdou, Ravishankar, Barrett, Belinkov, Elliott, and
  S{\o}gaard}]{DBLP:conf/acl/AbdouRBBES20}
Abdou, Mostafa, Vinit Ravishankar, Maria Barrett, Yonatan Belinkov, Desmond
  Elliott, and Anders S{\o}gaard. 2020.
\newblock The sensitivity of language models and humans to winograd schema
  perturbations.
\newblock In \emph{ACL}, pages 7590--7604, Association for Computational
  Linguistics.

\bibitem[{Alvarez-Melis and Jaakkola(2018)}]{alvarez2018towards}
Alvarez-Melis, David and Tommi~S Jaakkola. 2018.
\newblock Towards robust interpretability with self-explaining neural networks.
\newblock In \emph{NIPS}, pages 7786--7795.

\bibitem[{Alvarez{-}Melis and Jaakkola(2018)}]{DBLP:conf/nips/Alvarez-MelisJ18}
Alvarez{-}Melis, David and Tommi~S. Jaakkola. 2018.
\newblock Towards robust interpretability with self-explaining neural networks.
\newblock In \emph{NIPS}, pages 7786--7795.

\bibitem[{Arnold et~al.(2019)Arnold, Schneider, Cudré-Mauroux, Gers, and
  Löser}]{10.1162/tacl_a_00261}
Arnold, Sebastian, Rudolf Schneider, Philippe Cudré-Mauroux, Felix~A. Gers,
  and Alexander Löser. 2019.
\newblock {SECTOR: A Neural Model for Coherent Topic Segmentation and
  Classification}.
\newblock \emph{Transactions of the Association for Computational Linguistics},
  7:169--184.

\bibitem[{Bang et~al.(2021)Bang, Xie, Lee, Wu, and
  Xing}]{DBLP:conf/aaai/BangXL0X21}
Bang, Seo{-}Jin, Pengtao Xie, Heewook Lee, Wei Wu, and Eric~P. Xing. 2021.
\newblock Explaining {A} black-box by using {A} deep variational information
  bottleneck approach.
\newblock In \emph{Thirty-Fifth {AAAI} Conference on Artificial Intelligence,
  {AAAI} 2021, Thirty-Third Conference on Innovative Applications of Artificial
  Intelligence, {IAAI} 2021, The Eleventh Symposium on Educational Advances in
  Artificial Intelligence, {EAAI} 2021, Virtual Event, February 2-9, 2021},
  pages 11396--11404, {AAAI} Press.

\bibitem[{Bastings, Aziz, and Titov(2019)}]{DBLP:conf/acl/BastingsAT19}
Bastings, Jasmijn, Wilker Aziz, and Ivan Titov. 2019.
\newblock Interpretable neural predictions with differentiable binary
  variables.
\newblock In \emph{Proceedings of the 57th Conference of the Association for
  Computational Linguistics, {ACL} 2019, Florence, Italy, July 28- August 2,
  2019, Volume 1: Long Papers}, pages 2963--2977, Association for Computational
  Linguistics.

\bibitem[{Blei, Ng, and Jordan(2003)}]{blei2003latent}
Blei, David~M, Andrew~Y Ng, and Michael~I Jordan. 2003.
\newblock Latent dirichlet allocation.
\newblock \emph{JMLR}, 3:993--1022.

\bibitem[{Brown et~al.(2020)Brown, Mann, Ryder, Subbiah, Kaplan, Dhariwal,
  Neelakantan, Shyam, Sastry, Askell, Agarwal, Herbert{-}Voss, Krueger,
  Henighan, Child, Ramesh, Ziegler, Wu, Winter, Hesse, Chen, Sigler, Litwin,
  Gray, Chess, Clark, Berner, McCandlish, Radford, Sutskever, and
  Amodei}]{DBLP:conf/nips/BrownMRSKDNSSAA20}
Brown, Tom~B., Benjamin Mann, Nick Ryder, Melanie Subbiah, Jared Kaplan,
  Prafulla Dhariwal, Arvind Neelakantan, Pranav Shyam, Girish Sastry, Amanda
  Askell, Sandhini Agarwal, Ariel Herbert{-}Voss, Gretchen Krueger, Tom
  Henighan, Rewon Child, Aditya Ramesh, Daniel~M. Ziegler, Jeffrey Wu, Clemens
  Winter, Christopher Hesse, Mark Chen, Eric Sigler, Mateusz Litwin, Scott
  Gray, Benjamin Chess, Jack Clark, Christopher Berner, Sam McCandlish, Alec
  Radford, Ilya Sutskever, and Dario Amodei. 2020.
\newblock Language models are few-shot learners.
\newblock In \emph{NIPS}.

\bibitem[{Card, Tan, and Smith(2018)}]{DBLP:conf/acl/SmithCT18}
Card, Dallas, Chenhao Tan, and Noah~A. Smith. 2018.
\newblock Neural models for documents with metadata.
\newblock In \emph{ACL}, pages 2031--2040.

\bibitem[{Chaney and Blei(2021)}]{Chaney_Blei_2021}
Chaney, Allison and David Blei. 2021.
\newblock Visualizing topic models.
\newblock \emph{Proceedings of the International AAAI Conference on Web and
  Social Media}, 6(1):419--422.

\bibitem[{Chen and Ji(2020)}]{DBLP:conf/emnlp/ChenJ20}
Chen, Hanjie and Yangfeng Ji. 2020.
\newblock Learning variational word masks to improve the interpretability of
  neural text classifiers.
\newblock In \emph{EMNLP}, pages 4236--4251, Association for Computational
  Linguistics.

\bibitem[{Chen, Zheng, and Ji(2020)}]{DBLP:conf/acl/ChenZJ20}
Chen, Hanjie, Guangtao Zheng, and Yangfeng Ji. 2020.
\newblock Generating hierarchical explanations on text classification via
  feature interaction detection.
\newblock In \emph{Proceedings of the 58th Annual Meeting of the Association
  for Computational Linguistics, {ACL} 2020, Online, July 5-10, 2020}, pages
  5578--5593, Association for Computational Linguistics.

\bibitem[{Chen et~al.(2018)Chen, Song, Wainwright, and
  Jordan}]{DBLP:conf/icml/ChenSWJ18}
Chen, Jianbo, Le~Song, Martin~J. Wainwright, and Michael~I. Jordan. 2018.
\newblock Learning to explain: An information-theoretic perspective on model
  interpretation.
\newblock In \emph{ICML}, volume~80 of \emph{Proceedings of Machine Learning
  Research}, pages 882--891, {PMLR}.

\bibitem[{Chen et~al.(2020)Chen, Dai, Yuan, Lu, and
  Huang}]{DBLP:conf/acl/ChenDYLH20}
Chen, Jun, Xiaoya Dai, Quan Yuan, Chao Lu, and Haifeng Huang. 2020.
\newblock Towards interpretable clinical diagnosis with bayesian network
  ensembles stacked on entity-aware cnns.
\newblock In \emph{ACL}, pages 3143--3153, Association for Computational
  Linguistics.

\bibitem[{De-Arteaga et~al.(2019)De-Arteaga, Romanov, Wallach, Chayes, Borgs,
  Chouldechova, Geyik, Kenthapadi, and Kalai}]{de2019bias}
De-Arteaga, Maria, Alexey Romanov, Hanna Wallach, Jennifer Chayes, Christian
  Borgs, Alexandra Chouldechova, Sahin Geyik, Krishnaram Kenthapadi, and
  Adam~Tauman Kalai. 2019.
\newblock Bias in bios: A case study of semantic representation bias in a
  high-stakes setting.
\newblock In \emph{Proceedings of the Conference on Fairness, Accountability,
  and Transparency}, pages 120--128.

\bibitem[{Devlin et~al.(2019)Devlin, Chang, Lee, and
  Toutanova}]{DBLP:conf/naacl/DevlinCLT19}
Devlin, Jacob, Ming{-}Wei Chang, Kenton Lee, and Kristina Toutanova. 2019.
\newblock {BERT:} pre-training of deep bidirectional transformers for language
  understanding.
\newblock In \emph{NAACL}, pages 4171--4186.

\bibitem[{DeYoung et~al.(2020)DeYoung, Jain, Rajani, Lehman, Xiong, Socher, and
  Wallace}]{DBLP:conf/acl/DeYoungJRLXSW20}
DeYoung, Jay, Sarthak Jain, Nazneen~Fatema Rajani, Eric Lehman, Caiming Xiong,
  Richard Socher, and Byron~C. Wallace. 2020.
\newblock {ERASER:} {A} benchmark to evaluate rationalized {NLP} models.
\newblock In \emph{Proceedings of the 58th Annual Meeting of the Association
  for Computational Linguistics, {ACL} 2020, Online, July 5-10, 2020}, pages
  4443--4458, Association for Computational Linguistics.

\bibitem[{Guan et~al.(2019)Guan, Wang, Zhang, Chen, He, and
  Xie}]{DBLP:conf/icml/GuanWZCH019}
Guan, Chaoyu, Xiting Wang, Quanshi Zhang, Runjin Chen, Di~He, and Xing Xie.
  2019.
\newblock Towards a deep and unified understanding of deep neural models in
  {NLP}.
\newblock In \emph{ICML}, volume~97 of \emph{Proceedings of Machine Learning
  Research}, pages 2454--2463, {PMLR}.

\bibitem[{Gui and He(2021)}]{DBLP:journals/artmed/GuiH21}
Gui, Lin and Yulan He. 2021.
\newblock Understanding patient reviews with minimum supervision.
\newblock \emph{Artif. Intell. Medicine}, 120:102160.

\bibitem[{Gui et~al.(2022)Gui, Leng, Zhou, Xu, and
  He}]{DBLP:journals/tkde/GuiLZXH22}
Gui, Lin, Jia Leng, Jiyun Zhou, Ruifeng Xu, and Yulan He. 2022.
\newblock Multi task mutual learning for joint sentiment classification and
  topic detection.
\newblock \emph{{IEEE} Trans. Knowl. Data Eng.}, 34(4):1915--1927.

\bibitem[{Jacovi and Goldberg(2020)}]{DBLP:conf/acl/JacoviG20}
Jacovi, Alon and Yoav Goldberg. 2020.
\newblock Towards faithfully interpretable {NLP} systems: How should we define
  and evaluate faithfulness?
\newblock In \emph{ACL}, pages 4198--4205, Association for Computational
  Linguistics.

\bibitem[{Jain and Wallace(2019)}]{DBLP:conf/naacl/JainW19}
Jain, Sarthak and Byron~C. Wallace. 2019.
\newblock Attention is not explanation.
\newblock In \emph{Proceedings of the 2019 Conference of the North American
  Chapter of the Association for Computational Linguistics: Human Language
  Technologies, {NAACL-HLT} 2019, Minneapolis, MN, USA, June 2-7, 2019, Volume
  1 (Long and Short Papers)}, pages 3543--3556, Association for Computational
  Linguistics.

\bibitem[{Jawahar, Sagot, and Seddah(2019)}]{jawahar2019does}
Jawahar, Ganesh, Beno{\^\i}t Sagot, and Djam{\'e} Seddah. 2019.
\newblock What does bert learn about the structure of language?
\newblock In \emph{ACL}, pages 3651--3657.

\bibitem[{Jiang et~al.(2020)Jiang, Zhao, Chu, Shen, and
  Tu}]{DBLP:conf/emnlp/JiangZCST20}
Jiang, Chengyue, Yinggong Zhao, Shanbo Chu, Libin Shen, and Kewei Tu. 2020.
\newblock Cold-start and interpretability: Turning regular expressions into
  trainable recurrent neural networks.
\newblock In \emph{EMNLP}, pages 3193--3207, Association for Computational
  Linguistics.

\bibitem[{Jin et~al.(2020)Jin, Wei, Du, Xue, and
  Ren}]{DBLP:conf/iclr/JinWDXR20}
Jin, Xisen, Zhongyu Wei, Junyi Du, Xiangyang Xue, and Xiang Ren. 2020.
\newblock Towards hierarchical importance attribution: Explaining compositional
  semantics for neural sequence models.
\newblock In \emph{8th International Conference on Learning Representations,
  {ICLR} 2020, Addis Ababa, Ethiopia, April 26-30, 2020}, OpenReview.net.

\bibitem[{Johansson, Shalit, and Sontag(2016)}]{DBLP:conf/icml/JohanssonSS16}
Johansson, Fredrik~D., Uri Shalit, and David~A. Sontag. 2016.
\newblock Learning representations for counterfactual inference.
\newblock In \emph{ICML}, volume~48 of \emph{{JMLR} Workshop and Conference
  Proceedings}, pages 3020--3029, JMLR.org.

\bibitem[{Kim et~al.(2020)Kim, Yi, Kim, and Yoon}]{DBLP:conf/emnlp/KimYKY20}
Kim, Siwon, Jihun Yi, Eunji Kim, and Sungroh Yoon. 2020.
\newblock Interpretation of {NLP} models through input marginalization.
\newblock In \emph{EMNLP}, pages 3154--3167, Association for Computational
  Linguistics.

\bibitem[{Kingma and Ba(2015)}]{DBLP:journals/corr/KingmaB14}
Kingma, Diederik~P. and Jimmy Ba. 2015.
\newblock Adam: {A} method for stochastic optimization.
\newblock In \emph{3rd International Conference on Learning Representations,
  {ICLR} 2015, San Diego, CA, USA, May 7-9, 2015, Conference Track
  Proceedings}.

\bibitem[{Kingma and Welling(2014)}]{DBLP:journals/corr/KingmaW13}
Kingma, Diederik~P. and Max Welling. 2014.
\newblock Auto-encoding variational bayes.
\newblock In \emph{2nd International Conference on Learning Representations,
  {ICLR} 2014, Banff, AB, Canada, April 14-16, 2014, Conference Track
  Proceedings}.

\bibitem[{Lai and Tan(2019)}]{lai2019human}
Lai, Vivian and Chenhao Tan. 2019.
\newblock On human predictions with explanations and predictions of machine
  learning models: A case study on deception detection.
\newblock In \emph{Proceedings of the Conference on Fairness, Accountability,
  and Transparency}, pages 29--38.

\bibitem[{Lei, Barzilay, and Jaakkola(2016)}]{DBLP:conf/emnlp/LeiBJ16}
Lei, Tao, Regina Barzilay, and Tommi~S. Jaakkola. 2016.
\newblock Rationalizing neural predictions.
\newblock In \emph{EMNLP}, pages 107--117, The Association for Computational
  Linguistics.

\bibitem[{Li, Monroe, and Jurafsky(2016)}]{DBLP:journals/corr/LiMJ16a}
Li, Jiwei, Will Monroe, and Dan Jurafsky. 2016.
\newblock Understanding neural networks through representation erasure.
\newblock \emph{CoRR}, abs/1612.08220.

\bibitem[{Li and Eisner(2019)}]{DBLP:conf/emnlp/LiE19}
Li, Xiang~Lisa and Jason Eisner. 2019.
\newblock Specializing word embeddings (for parsing) by information bottleneck.
\newblock In \emph{Proceedings of the 2019 Conference on Empirical Methods in
  Natural Language Processing and the 9th International Joint Conference on
  Natural Language Processing, {EMNLP-IJCNLP} 2019, Hong Kong, China, November
  3-7, 2019}, pages 2744--2754, Association for Computational Linguistics.

\bibitem[{Lin et~al.(2012)Lin, He, Everson, and Ruger}]{5710933}
Lin, Chenghua, Yulan He, Richard Everson, and Stefan Ruger. 2012.
\newblock Weakly supervised joint sentiment-topic detection from text.
\newblock \emph{IEEE Transactions on Knowledge and Data Engineering},
  24(6):1134--1145.

\bibitem[{Lipton(2018)}]{DBLP:journals/cacm/Lipton18}
Lipton, Zachary~C. 2018.
\newblock The mythos of model interpretability.
\newblock \emph{Commun. {ACM}}, 61(10):36--43.

\bibitem[{Maas et~al.(2011)Maas, Daly, Pham, Huang, Ng, and
  Potts}]{maas2011learning}
Maas, Andrew, Raymond~E Daly, Peter~T Pham, Dan Huang, Andrew~Y Ng, and
  Christopher Potts. 2011.
\newblock Learning word vectors for sentiment analysis.
\newblock In \emph{ACL-HLT}, pages 142--150.

\bibitem[{McAuley, Leskovec, and Jurafsky(2012)}]{mcauley2012learning}
McAuley, Julian, Jure Leskovec, and Dan Jurafsky. 2012.
\newblock Learning attitudes and attributes from multi-aspect reviews.
\newblock In \emph{2012 IEEE 12th International Conference on Data Mining},
  pages 1020--1025, IEEE.

\bibitem[{Niu et~al.(2020)Niu, Mathur, Dinu, and
  Al{-}Onaizan}]{DBLP:conf/acl/NiuMDA20}
Niu, Xing, Prashant Mathur, Georgiana Dinu, and Yaser Al{-}Onaizan. 2020.
\newblock Evaluating robustness to input perturbations for neural machine
  translation.
\newblock In \emph{ACL}, pages 8538--8544, Association for Computational
  Linguistics.

\bibitem[{O'Hare et~al.(2009)O'Hare, Davy, Bermingham, Ferguson, Sheridan,
  Gurrin, and Smeaton}]{DBLP:conf/cikm/OHareDBFSGS09}
O'Hare, Neil, Michael Davy, Adam Bermingham, Paul Ferguson, P{\'{a}}raic
  Sheridan, Cathal Gurrin, and Alan~F. Smeaton. 2009.
\newblock Topic-dependent sentiment analysis of financial blogs.
\newblock In \emph{Proceedings of the 1st International {CIKM} Workshop on
  Topic-Sentiment Analysis for Mass Opinion, {TSA} '09, Hong Kong, SAR, China,
  November 6, 2009}, pages 9--16, {ACM}.

\bibitem[{Pruthi et~al.(2020)Pruthi, Gupta, Dhingra, Neubig, and
  Lipton}]{DBLP:conf/acl/PruthiGDNL20}
Pruthi, Danish, Mansi Gupta, Bhuwan Dhingra, Graham Neubig, and Zachary~C.
  Lipton. 2020.
\newblock Learning to deceive with attention-based explanations.
\newblock In \emph{Proceedings of the 58th Annual Meeting of the Association
  for Computational Linguistics, {ACL} 2020, Online, July 5-10, 2020}, pages
  4782--4793, Association for Computational Linguistics.

\bibitem[{Ribeiro et~al.(2020)Ribeiro, Wu, Guestrin, and
  Singh}]{DBLP:conf/acl/RibeiroWGS20}
Ribeiro, Marco~T{\'{u}}lio, Tongshuang Wu, Carlos Guestrin, and Sameer Singh.
  2020.
\newblock Beyond accuracy: Behavioral testing of {NLP} models with checklist.
\newblock In \emph{ACL}, pages 4902--4912, Association for Computational
  Linguistics.

\bibitem[{Rieger et~al.(2020)Rieger, Singh, Murdoch, and
  Yu}]{rieger2020interpretations}
Rieger, Laura, Chandan Singh, William Murdoch, and Bin Yu. 2020.
\newblock Interpretations are useful: penalizing explanations to align neural
  networks with prior knowledge.
\newblock In \emph{ICML}, pages 8116--8126.

\bibitem[{Selvaraju et~al.(2020)Selvaraju, Cogswell, Das, Vedantam, Parikh, and
  Batra}]{DBLP:journals/ijcv/SelvarajuCDVPB20}
Selvaraju, Ramprasaath~R., Michael Cogswell, Abhishek Das, Ramakrishna
  Vedantam, Devi Parikh, and Dhruv Batra. 2020.
\newblock Grad-cam: Visual explanations from deep networks via gradient-based
  localization.
\newblock \emph{IJCV}, 128(2):336--359.

\bibitem[{Serrano and Smith(2019)}]{DBLP:conf/acl/SerranoS19}
Serrano, Sofia and Noah~A. Smith. 2019.
\newblock Is attention interpretable?
\newblock In \emph{Proceedings of the 57th Conference of the Association for
  Computational Linguistics, {ACL} 2019, Florence, Italy, July 28- August 2,
  2019, Volume 1: Long Papers}, pages 2931--2951, Association for Computational
  Linguistics.

\bibitem[{Singh, Murdoch, and Yu(2019)}]{singh2019hierarchical}
Singh, Chandan, W~James Murdoch, and Bin Yu. 2019.
\newblock Hierarchical interpretations for neural network predictions.
\newblock In \emph{ICLR}.

\bibitem[{Tang, Hahn{-}Powell, and Surdeanu(2020)}]{DBLP:conf/acl/TangHS20}
Tang, Zheng, Gus Hahn{-}Powell, and Mihai Surdeanu. 2020.
\newblock Exploring interpretability in event extraction: Multitask learning of
  a neural event classifier and an explanation decoder.
\newblock In \emph{ACL}, pages 169--175, Association for Computational
  Linguistics.

\bibitem[{Wang et~al.(2020)Wang, Wang, Zhang, Duan, Zhou, and
  Chen}]{pmlr-v108-wang20l}
Wang, Zhengjue, Chaojie Wang, Hao Zhang, Zhibin Duan, Mingyuan Zhou, and
  Bo~Chen. 2020.
\newblock Learning dynamic hierarchical topic graph with graph convolutional
  network for document classification.
\newblock In \emph{Proceedings of the Twenty Third International Conference on
  Artificial Intelligence and Statistics}, volume 108 of \emph{Proceedings of
  Machine Learning Research}, pages 3959--3969, PMLR.

\bibitem[{Wiegreffe and Pinter(2019)}]{DBLP:conf/emnlp/WiegreffeP19}
Wiegreffe, Sarah and Yuval Pinter. 2019.
\newblock Attention is not not explanation.
\newblock In \emph{Proceedings of the 2019 Conference on Empirical Methods in
  Natural Language Processing and the 9th International Joint Conference on
  Natural Language Processing, {EMNLP-IJCNLP} 2019, Hong Kong, China, November
  3-7, 2019}, pages 11--20, Association for Computational Linguistics.

\bibitem[{Wu et~al.(2020)Wu, Chen, Kao, and Liu}]{DBLP:conf/acl/WuCKL20}
Wu, Zhiyong, Yun Chen, Ben Kao, and Qun Liu. 2020.
\newblock Perturbed masking: Parameter-free probing for analyzing and
  interpreting {BERT}.
\newblock In \emph{ACL}, pages 4166--4176, Association for Computational
  Linguistics.

\bibitem[{Xie et~al.(2021)Xie, Huang, Du, Peng, and
  Nie}]{10.1145/3442381.3450045}
Xie, Qianqian, Jimin Huang, Pan Du, Min Peng, and Jian-Yun Nie. 2021.
\newblock Graph topic neural network for document representation.
\newblock In \emph{Proceedings of the Web Conference 2021}, WWW '21, page
  3055–3065, Association for Computing Machinery, New York, NY, USA.

\bibitem[{Yan et~al.(2021)Yan, Gui, Pergola, and He}]{DBLP:conf/acl/Yan0PH20}
Yan, Hanqi, Lin Gui, Gabriele Pergola, and Yulan He. 2021.
\newblock Position bias mitigation: {A} knowledge-aware graph model for emotion
  cause extraction.
\newblock In \emph{Proceedings of the 59th Annual Meeting of the Association
  for Computational Linguistics and the 11th International Joint Conference on
  Natural Language Processing, {ACL/IJCNLP} 2021, (Volume 1: Long Papers),
  Virtual Event, August 1-6, 2021}, pages 3364--3375, Association for
  Computational Linguistics.

\bibitem[{Yang et~al.(2019)Yang, Dai, Yang, Carbonell, Salakhutdinov, and
  Le}]{DBLP:conf/nips/YangDYCSL19}
Yang, Zhilin, Zihang Dai, Yiming Yang, Jaime~G. Carbonell, Ruslan
  Salakhutdinov, and Quoc~V. Le. 2019.
\newblock Xlnet: Generalized autoregressive pretraining for language
  understanding.
\newblock In \emph{NIPS}, pages 5754--5764.

\bibitem[{Yang et~al.(2016)Yang, Yang, Dyer, He, Smola, and
  Hovy}]{DBLP:conf/naacl/YangYDHSH16}
Yang, Zichao, Diyi Yang, Chris Dyer, Xiaodong He, Alexander~J. Smola, and
  Eduard~H. Hovy. 2016.
\newblock Hierarchical attention networks for document classification.
\newblock In \emph{NAACL}, pages 1480--1489, The Association for Computational
  Linguistics.

\bibitem[{Yu et~al.(2021)Yu, Zhang, Chang, and
  Jaakkola}]{DBLP:conf/nips/YuZCJ21}
Yu, Mo, Yang Zhang, Shiyu Chang, and Tommi~S. Jaakkola. 2021.
\newblock Understanding interlocking dynamics of cooperative rationalization.
\newblock In \emph{Advances in Neural Information Processing Systems 34: Annual
  Conference on Neural Information Processing Systems 2021, NeurIPS 2021,
  December 6-14, 2021, virtual}, pages 12822--12835.

\bibitem[{Zanzotto et~al.(2020)Zanzotto, Santilli, Ranaldi, Onorati, Tommasino,
  and Fallucchi}]{DBLP:conf/emnlp/ZanzottoSROTF20}
Zanzotto, Fabio~Massimo, Andrea Santilli, Leonardo Ranaldi, Dario Onorati,
  Pierfrancesco Tommasino, and Francesca Fallucchi. 2020.
\newblock {KERMIT:} complementing transformer architectures with encoders of
  explicit syntactic interpretations.
\newblock In \emph{EMNLP}, pages 256--267, Association for Computational
  Linguistics.

\bibitem[{Zhang et~al.(2020)Zhang, Sun, Feng, and
  Li}]{DBLP:conf/acl/ZhangSFL20}
Zhang, Jingyuan, Mingming Sun, Yue Feng, and Ping Li. 2020.
\newblock Learning interpretable relationships between entities, relations and
  concepts via bayesian structure learning on open domain facts.
\newblock In \emph{ACL}, pages 8045--8056, Association for Computational
  Linguistics.

\bibitem[{Zhou, Zhang, and Yang(2020)}]{DBLP:conf/acl/ZhouZY20}
Zhou, Fan, Shengming Zhang, and Yi~Yang. 2020.
\newblock Interpretable operational risk classification with semi-supervised
  variational autoencoder.
\newblock In \emph{ACL}, pages 846--852, Association for Computational
  Linguistics.

\bibitem[{Zhou et~al.(2020)Zhou, Hu, Zhang, Liang, Sun, Xiong, and
  Tang}]{DBLP:conf/nips/ZhouHZLSXT20}
Zhou, Wangchunshu, Jinyi Hu, Hanlin Zhang, Xiaodan Liang, Maosong Sun, Chenyan
  Xiong, and Jian Tang. 2020.
\newblock Towards interpretable natural language understanding with
  explanations as latent variables.
\newblock In \emph{Advances in Neural Information Processing Systems 33: Annual
  Conference on Neural Information Processing Systems 2020, NeurIPS 2020,
  December 6-12, 2020, virtual}, pages 6803–--6814.

\end{thebibliography}
\clearpage
\onecolumnnew

\appendix


\setcounter{table}{0}
\makeatletter 
\renewcommand{\thetable}{A\@arabic\c@table}
\makeatother

\appendixsection{Human Evaluation Instruction}
\label{appendix:humanevaluate}

\hqDec{\textsc{Hint}, HAN, VAMSK generate different forms of interpretations. HAN can generate the interpretations based on the attention weights at both the word-level and the sentence-level. VMASK can only generate the interpretations at the word-level. Apart from the word-level and the sentence-level interpretations, \textsc{Hint} can also generate interpretations at the document-level by partitioning sentences into various topics and associating with each topic a polarity label. In the actual evaluation, to reduce cognitive load, we only present the most prominent topic in the document and the most contrastive topic in the form of word clouds to the evaluators. To retrieve the most prominent topic, we first identify the topic dimension with the largest value in the latent topic vector for each sentence and then select the most common topic dimension among all sentences. 
To select the most contrastive topic, we choose the one which has the minimal similarity with the first chosen topic. To generate the word cloud, we retrieve the topic words following the approach discussed in Section~\ref{sec:generate_interpretation} with the vocabulary constrained to the local document. The evaluation schema is shown below:} 




\begin{tcolorbox}[center={y shift=-3mm,yshifttext=-1mm},colframe=yellow!50!black,colbacktitle=yellow!75!black,title= Evaluation Schema,fonttitle=\bfseries]
\tcblower
\begin{itemize}
\item \textbf{Correctness} $-$ it measures to what extend the model-generated interpretation could lead to correct prediction. We present the interpretations generated by a model, and ask an evaluator to predict the document label based solely on the interpretations and check if the predicted label agrees with the \textbf{ground-truth label}.
\item \textbf{Faithfulness} $-$ it measures to what extend the interpretation generated is faithful to the model prediction. The evaluators should check if the interpretation generated will lead to the \textbf{model predicted label}.
\item \textbf{Informativeness} $-$ it measures to what extend the interpretation reveals the key information conveyed in text. We present the identified important words from HAN and VMASK; sentence importance scores from HAN and \textsc{Hint}. Additionally, we present the topic word clouds from \textsc{Hint}. 
We then ask users to evaluate the following aspects:
\begin{itemize}
\item I know what the main topic is;
 \item I can easily tell the polarity of the main topic;
 \item I know what the secondary topic is (if there is any).
\end{itemize}
We use the 1-5 likert scale (\textbf{strongly disagree, disagree, neutral, agree, strongly agree}) for each of the criteria above.
\end{itemize}
\end{tcolorbox}



\appendixsection{List of keywords used for Yelp reviews retrieval}

The list of keywords used for retrieving patient reviews from Yelp is shown in Table \ref{tab:keywords}.

\begin{table}[!t]
    \centering
    \caption{ Keywords used to retrieve patient reviews from Yelp.}\label{tab:keywords}
    \small
    \begin{tabular}{p{13cm}} \toprule
    Walk-in Clinics,
    Surgeons,
    Oncologist,
    Cardiologists,
    Hospitals,
    Internal Medicine,
    Assisted Living Facilities,
    Cannabis Dispensaries,
    Doctors,
    Home Health Care,
    Health Coach,
    Emergency Pet Hospital,
    Pharmacy,
    Sleep Specialists,
    Professional Services,
    Addiction Medicine,
    Weight Loss Centers,
    Pediatric Dentists,
    Cosmetic Surgeons,
    Nephrologists,
    Naturopathic\/Holistic,
    Pediatricians,
    Nurse Practitioner,
    Urgent Care,
    Orthopedists,
    Drugstores,
    Optometrists,
    Rehabilitation Center,
    Hypnosis\/Hypnotherapy,
    Physical Therapy,
    Neurologist,
    Memory Care,
    Allergists,
    Counseling \& Mental Health,
    Pet Groomers,
    Podiatrists,
    Dermatologists,
    Diagnostic Services,
    Radiologists,
    Medical Centers,
    Gastroenterologist,
    Obstetricians \& Gynecologists,
    Pulmonologist,
    Ear Nose \& Throat,
    Ophthalmologists,
    Sports Medicine,
    Nutritionists,
    Psychiatrists,
    Vascular Medicine,
    Cannabis Clinics,
    Hospice,
    First Aid Classes,
    Medical Spas,
    Spine Surgeons,
    Health Retreats,
    Medical Transportation,
    Dentists,
    Health \& Medical,
    Speech Therapists,
    Emergency Medicine,
    Chiropractors,
    Medical Supplies,
    General Dentistry,
    Occupational Therapy,
    Urologists
    \\ \bottomrule
    \end{tabular}
    \end{table}


\appendixsection{Model Architecture and Parameter Setting}

\setcounter{table}{1}

\begin{table}[htb]
\centering
    \caption{ Model Architecture.}\small
    \begin{tabular}{p{2cm}p{3cm}p{7cm}}
    \toprule[1pt]
\textbf{Input}:&\multicolumn{2}{l}{A document $d$ consists of $M_d$ sentences $\{s_{i}\}_{i=1}^{M_d}$, $s_{i}=\{x_{ij}\}_{j=1}^{L}$}  \\ \hline

\textbf{Word Emb}&\multicolumn{2}{l}{ Initialised by the GloVe embedding,  $\{\bm{x}_{ij}\}_{j=1}^{L}\in \mathbb{R}^{N\times L}$  }   \\ \hline
\multirow{3}{*}{\textbf{Context learn}}& 
Word-level biLSTM& $\{\bm{x}_{ij}\}_{j=1}^{L}-\{\text{biLSTM}\}\rightarrow$: $\{\bm{h}_{ij}\}_{j=1}^{L}\in \mathbb{R}^{N\times L}$ \\

&Attention layer&$\{\bm{h}_{ij}\}_{j=1}^{L}-\{\rm{Linear}_{1}\}-\{\text{Linear}_{2}\}\rightarrow {\alpha_{ij}}\in \mathbb{R}^{L}$ \\
&Context aggregate&$\sum_{j=1}^{L}\{\bm{h}_{ij}\}{\alpha_{ij}}\rightarrow \bm{s}_{i}\in \mathbb{R}^{N}$.\\ \hline

\multirow{10}{*}{\textbf{Topic learn}} 
& Word Weight Init.& $\bm{x}_{i} = \sum_{j=1}^{L}\text{TFIDF}_{ij}$ $\cdot\bm{x}_{ij}$\\ \cline{2-3}

&\multirow{5}{*}{Bayesian inference}&
$\bm{x}_{i}-\{\rm{Encoder_{1}}\}\rightarrow\bm{\mu}_{\omega}\in \mathbb{R}^{N}$ \\
&&$\bm{x}_{i}-\{\rm{Encoder_{2}}\}\rightarrow \text{log}{\bm{\sigma}^{2}_{\omega}}\in \mathbb{R}^{N}$\\
&&$\bm{\omega}= \rm{Softmax}(\mu_{\omega} + \sigma_{\omega} \cdot \epsilon)$, $\epsilon  \sim \mathcal{N}(0,\mathbf{I})$
\\ \cline{2-3}

&\multirow{4}{*}{Autorenoder} & 
$\bm{\beta}^{1\times L}= \text{softmax}(\text{ReLU}(\bm{\omega}^{1 \times d}\cdot\bm{x}_{i})$\\
&& $\bm{r}_i = \sum^L_{j=1} \beta_{ij} \cdot \bm{x}_{ij}$\\
&& $\bm{z}_i = {\rm softmax}(W_c \cdot \bm{x}_i + b_c)$\\ 
&&$\bm{r}'_i = {\rm tanh}(W'_c \cdot \bm{z}_i + b'_c)$\\
\hline
\multirow{3}{*}{\textbf{Doc Modeling}}& Node Init.& 

$\bm{s}_{i}-\{\rm{Linear3}\}\rightarrow\{\rm{Linear4}\}\rightarrow$ $\bm{s}_{i}^{0}$\\
&Edge weight init.& $e_{ij}=\rm{softmax}(\bm{z}_{i}^\intercal \bm{z}_{j})$\\
&Node update&$\bm{s}_{i}^{\ell+1} = \sigma(\sum_{j\in \mathcal{N}_{i}}e_{ij}\mathbf{W}s_{j}^{\ell})$\\
&&$\bm{d} = (\textbf{s}_{1}^{L}+\textbf{s}_{2}^{L}...+\textbf{s}_{M_d}^{L})/M_{d}$\\
\hline
\textbf{Classification}&\multicolumn{2}{l}{$\hat{y}= \text{softmax}\Big(\rm{Linear}_{6}\big(\rm{LeakyReLU}(\rm{Linear}_{5}(\bm{d}))\big)\Big)$}\\
\bottomrule
    \end{tabular}
    \label{tab:model_structure}
\end{table}

Our model architecture is shown in Table~\ref{tab:model_structure}. We describe the parameter setup for each part of the model below:

\begin{itemize}
\item\emph{Context Learning} We use the pretrained 300-dimension GloVe embeddings with the dimension $N=300$. The dimension of the word-level biLSTM hidden states is $150$, and the dimension of the output $\bm{x}$ is also $300$. The word embedding sequence is fed to two consecutive linear layers to obtain the attention weights. The weight matrices for the linear layers, $\rm{Linear}_{1}$ and  $\rm{Linear}_{2}$, are $(300,200)$ and $(200,1)$, respectively. Then we aggregate $\bm{\hat{x}}$ by attention weights to obtain the sentence-level contextual representation $\bm{s}_{i}$.

\item\emph{Topic Learning} We calculate the TFIDF values for words offline. During inference, the TFIDF value of the out-of-vocabulary words is set to $1e-4$. For each sentence, we first normalise the TFIDF values of its constituent words and then aggregate the word embeddings weighted by their respective TFIDF values. This gives an initial sentence representation $\bm{x}_{i}\in \mathbb{R}^{N}$, which is then fed into two MLPs to generate the mean $\bm{\mu}_{\omega}$ and the variance $\log{\sigma}^{2}_{\omega}$. The output latent variable $\bm{\omega}\in \mathbb{R}^{N}$. After non-linear (\texttt{ReLU}) transformation and normalisation (\texttt{Softmax}), we obtain the topic-aware weights $\bm{\beta}$ that is used to generate the input $p_{i}$ for the autoenoder. The encoder and decoder in our autoencoder are 1-layer MLP with non-linear transformation. The weight matrices $\bm{W}_{c}$ and $\bm{W}_{c}'$ are $(300,K)$ and $(K,300)$ respectively. $K$ is the number of pre-defined topics. We set $K=50$ for the two review datasets and $K=30$ for the Guardian News data empirically.

\item\emph{Document Modeling} Graph nodes are initialised by the linear-transformed contextual sentence-level representations. The weight matrix in $\rm{Linear}_{3}$ and $\rm{Linear}_{4}$ are $(300, 200)$ and $(200,50)$. The graph node dimension is 50.

\item\emph{Classification} The $\rm{Linear}_{5}$ and $\rm{Linear}_{6}$ have the dimensions of $(300,200)$ and $(200,\text{\#labels})$, respectively.
\end{itemize}

We use dropout layers to alleviate over-fitting, and insert a dropout layer after the word embedding layer, the word-level biLSTM layer, and after obtaining $\bm{z}_{i}$ and $\bm{\omega}$, respectively. The dropout rate is $0.4$. We use the Adam~\cite{DBLP:journals/corr/KingmaB14} optimiser and set the learning rate to $1e-4$. The  $\lambda_{1}$ and $\lambda_{2}$ in the regularisation term are set to $0.05$ and $0.01$, respectively. $\eta_{a}$ and $\eta_{b}$ are set to 0.001 and 1, respectively. We train the model for $30$ epochs and evaluate the performance at the end of each epoch. We report the average results for running $5$ times with random seeds.
\end{document}